\documentclass[10pt,journal,compsoc]{IEEEtran}
\usepackage{amsmath,amsfonts}
\usepackage{algorithmic}
\usepackage{algorithm}
\usepackage{amssymb}
\usepackage{array}
\usepackage{textcomp}
\usepackage{stfloats}
\usepackage{url}
\usepackage{verbatim}
\usepackage{graphicx}
\usepackage{cite}
\usepackage{color}
\usepackage{booktabs}
\usepackage{threeparttable}
\usepackage{bbding}
\usepackage{pifont}
\usepackage{multicol}
\usepackage[caption=false,font=footnotesize,labelfont=rm,textfont=rm]{subfig}

\ifodd 1
\newcommand{\com}[1]{\textbf{\color{red}(COMMENT: #1)}} 
\else

\newcommand{\com}[1]{}
\fi

\newcommand{\revise}[1]{{\color{black}{#1}}}

\newcommand{\majorrevise}[1]{{\color{black}{#1}}}
\newcommand{\waittoadd}[1]{{\color{black}{#1}}}

\def\eg{e.g.}
\def\ie{i.e.}

\ifCLASSINFOpdf

\else

\fi

\usepackage{balance}

\begin{document}

\title{\fontsize{19pt}{\baselineskip}\selectfont { Flight Dynamics to Sensing Modalities: Exploiting Drone Ground Effect for Accurate Edge Detection }}

\author{
Chenyu Zhao, Jingao Xu, Ciyu Ruan, Haoyang Wang, Shengbo Wang, Jiaqi Li, Jirong Zha, Weijie Hong, 
Zheng Yang, ~\IEEEmembership{Fellow, ~IEEE},
Yunhao Liu, ~\IEEEmembership{Fellow, ~IEEE},
Xiao-Ping Zhang, ~\IEEEmembership{Fellow, ~IEEE},
Xinlei Chen, ~\IEEEmembership{Member, ~IEEE}

\IEEEcompsocitemizethanks{
\IEEEcompsocthanksitem A preliminary version of this article appeared in ACM Mobile Computing and Networking (ACM MobiCom2024) \cite{wang2024transformloc}.

\IEEEcompsocthanksitem Chenyu Zhao, Ciyu Ruan, Haoyang Wang, Shengbo Wang, Jiaqi Li, and Jirong Zha are with Shenzhen International Graduate School, Tsinghua University, China. \\
E-mail: \{zhaocyhi, softword77, wanghaoyang0428, shengbo1.wang, jackielithu, zhajirong\}@gmail.com

\IEEEcompsocthanksitem Jingao Xu is with the Computer Science Department, Carnegie Mellon University, USA. \\
E-mail: jingaox@andrew.cmu.edu

\IEEEcompsocthanksitem Weijie Hong is with Shenzhen Smart City Communication Co., Ltd., China. \\
E-mail: hongweijie@smartcitysz.com

\IEEEcompsocthanksitem Zheng Yang and Yunhao Liu are with the School of Software, Tsinghua University, Beijing 100084, China. \\
Email: \{hmilyyz, yunhaoliu\}@gmail.com

\IEEEcompsocthanksitem Xiao-Ping Zhang is with Shenzhen International Graduate School, Tsinghua University, China.\\
E-mail: xpzhang@ieee.org

\IEEEcompsocthanksitem Xinlei Chen is with Shenzhen International Graduate School, Tsinghua University, China.\\
E-mail: chen.xinlei@sz.tsinghua.edu.cn

\IEEEcompsocthanksitem Co-primary authors: Chenyu Zhao, Jingao Xu, and Ciyu Ruan
\IEEEcompsocthanksitem Corresponding author: Xinlei Chen.
\IEEEcompsocthanksitem Manuscript submitted January 2025.

}
}

\markboth{IEEE TRANSACTIONS ON MOBILE COMPUTING}%
{Wang. H \MakeLowercase{\textit{et al.}}: TransformLoc}

\IEEEtitleabstractindextext{%
\begin{abstract}

\revise{Drone-based rapid and accurate environmental edge detection is highly advantageous for tasks such as disaster relief and autonomous navigation. \majorrevise{Current methods, using radars or cameras, raise deployment costs and burden lightweight drones with high computational demands.} In this paper, we propose AirTouch, a system that transforms the ground effect from a stability "foe" in traditional flight control views, into a "friend" for accurate and efficient edge detection. \majorrevise{Our key insight is that analyzing drone basic attitude sensor readings and flight commands allows us to detect ground effect changes. Such changes typically indicate the drone flying over a boundary of two materials, making this information valuable for edge detection. We approach this insight through theoretical analysis, algorithm design, and implementation, fully leveraging the ground effect as a new sensing modality without compromising drone flight stability, thereby achieving accurate and efficient scene edge detection. We also compare this new sensing modality with vision-based methods to clarify its exclusive advantages in resource efficiency and detection capability. Extensive evaluations demonstrate that our system achieves a high detection accuracy with mean detection distance errors of 0.051m, outperforming the baseline method performance by $86\%$. With such detection performance, our system requires only 43 mW power consumption, contributing to this new sensing modality for low-cost and highly efficient edge detection.}}

\end{abstract}

}

\maketitle

\IEEEdisplaynontitleabstractindextext

\IEEEpeerreviewmaketitle

\vspace{-2cm}
\IEEEraisesectionheading{\section{Introduction}}

\IEEEPARstart{R}{apid} and accurate environmental edge detection is crucial in various applications, including disaster response~\cite{sie2023batmobility, chen2024soscheduler, chen2020pas}, rescue-and-relief~\cite{zhang2023rf, hsia2024demo}, and autonomous navigation~\cite{lu2020see, lu2020milliego, liu2024mobiair}. A key aspect involves detecting terrain edges, such as sudden changes in height (\eg, steps, cliffs) and variations in ground materials (\eg, water, soil, solid rock). With prior knowledge of these edges, intelligent systems can plan paths for humans and robots more logically, efficiently, and safely ~\cite{chung2018survey, mur2017orb, li2024edgeslam2, xu2019ilocus}. To boost efficiency and cut costs in large-scale edge detection~\cite{li2022tract}, mainstream systems leverage swarms of lightweight drones~\cite{chen2017design} (a.k.a., UAV) to execute the task collaboratively as they fly and scan the entire scene~\cite{chen2015drunkwalk, xu2022swarmmap, wang2023TransformLoc, chen2020h, chen2024ddl}.

\waittoadd{
In disaster relief scenarios, such as earthquake rescue or flood response, the ability to quickly and precisely detect edges, including sudden height changes and material boundaries, can significantly enhance the efficiency and safety of rescue missions~\cite{dai2020integrated, ijaz2023uav}. For example, drones equipped with edge detection capabilities can navigate through debris fields, identify safe landing zones, and locate survivors trapped in complex terrains. This enables rescue teams to reach affected areas more rapidly and with greater precision, potentially saving lives~\cite{9740826, tang2024review}. In autonomous navigation, edge detection plays a vital role in enabling drones to operate safely and efficiently in unknown environments. By detecting edges such as cliffs, steps, and material transitions, drones can avoid obstacles, plan optimal flight paths, and make real-time decisions to ensure smooth and collision-free navigation~\cite{kan2013extreme, gyagenda2022review}. This is particularly important in scenarios where GPS signals are weak or unavailable, such as in indoor environments or dense forests. Accurate edge detection allows drones to maintain situational awareness and adapt to dynamic surroundings, improving their autonomy and reliability.

}

\waittoadd{
Existing drone-based edge detection methods face challenges, categorized into two main types:
(\textit{i}) Wireless-signal based methods: These methods utilize signals like mmWave Radar~\cite{sie2023batmobility}, Wi-Fi~\cite{hu2023wisdom}, terahertz radar~\cite{afzal2023agritera}, LiDAR~\cite{li2023leovr}, and acoustic signals~\cite{wang2023meta} for contactless edge detection. 
While effective, they often rely on infrastructure and struggle with operations in inaccessible areas~\cite{wang2022micnest}, and are typically power-hungry, making them less suitable for lightweight drones with limited battery capacity.
(\textit{ii}) Visual-sensor-based methods: These algorithms aim to accurately detect object boundaries using computer vision techniques or neural networks~\cite{he2020bdcn, huan2021unmixing, soria2023dense}. 
However, they require significant computational resources, limiting their deployment on resource-constrained drones~\cite{liu2020dynamic}, which can be a bottleneck for real-time applications. Visual sensors like cameras struggle in low-light conditions or when dealing with surfaces of similar colors and textures, which can compromise their detection accuracy and robustness.
Furthermore, the existing pioneer study~\cite{wang2024multi} proposes multi-modal fusion sensing, achieving promising accuracy by integrating the information from both wireless and visual domains.
}


\begin{figure}[t]
\centering
\includegraphics[width=0.95\linewidth]{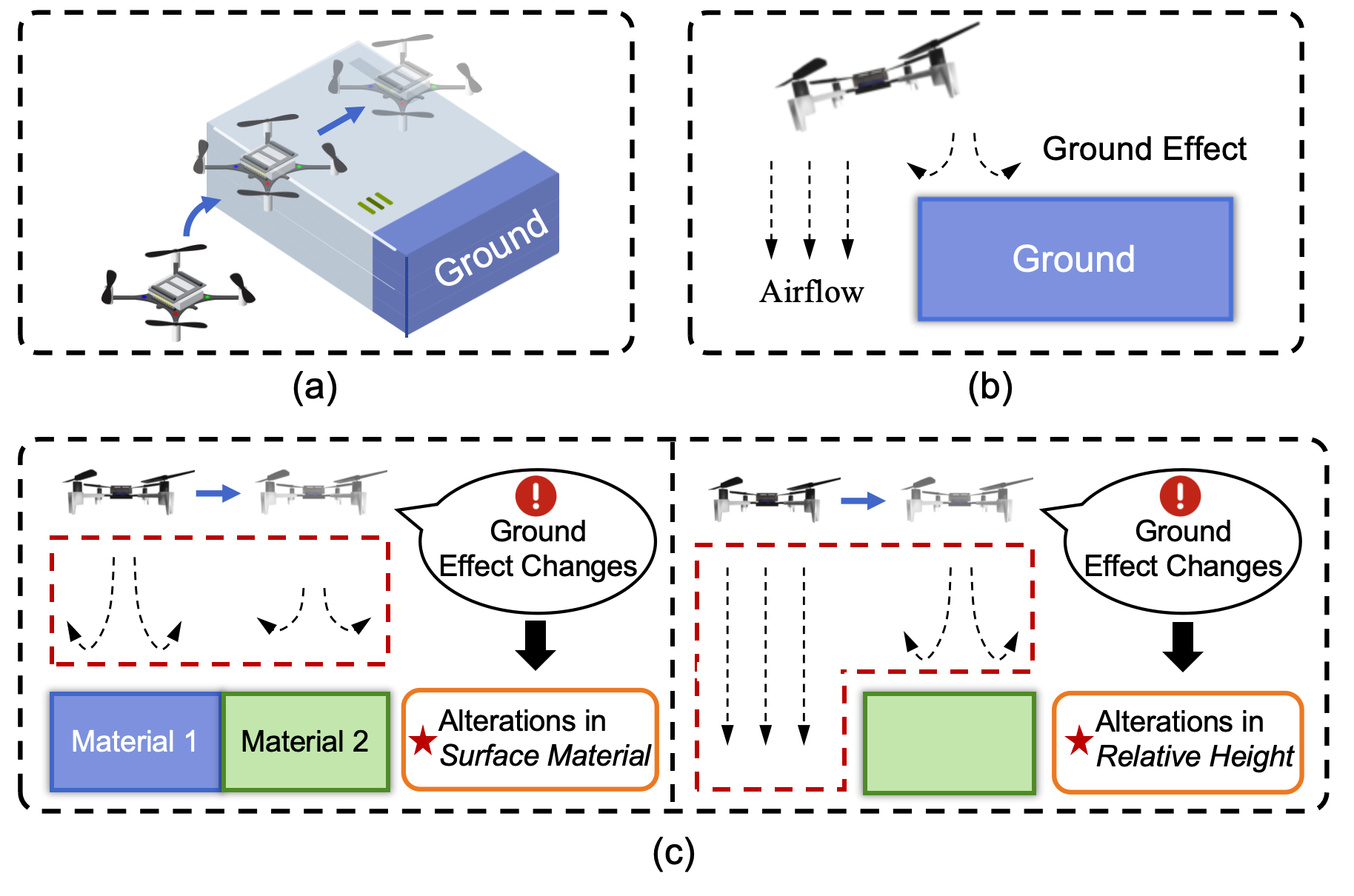}
\vspace{-0.8em}
\caption{\majorrevise{(a) A drone utilizes ground effect for edge detection. (b) When a drone flies closely to the ground, the increased upward lift is caused by the \textbf{ground effect}. (c) By detecting variations in ground effect, we can infer abrupt changes in height or alterations in surface materials, thereby identifying scene boundaries.}}
\vspace{-1.5em}
\label{scenario}
\end{figure}

In this work, we aim to introduce a novel approach for edge detection on lightweight drones, which will serve as a significant complement to the aforementioned two envelopes, especially in situations where computational resources are scarce and scene visibility is low.
\majorrevise{As illustrated in Fig~.\ref{scenario}, when drones fly closely to the ground (\ie, a surface): the airflow generated by the rotating rotors bounces off the surface below the drone, creating additional upward lift and leading to disturbances in a drone's flight state.} 
This phenomenon, widely known as \textbf{ground effect} (GE)~\cite{conyers2018empirical, zhao2024demo}, varies with the drone's altitude above the surface and the type of surface material.

\waittoadd{\textbf{Ground effect of drones.}
Achieving precise control over drone positions is paramount, yet remains a considerable challenge \cite{bai2022bioinspired, song2023reaching}. 
This challenge is predominantly attributed to the intricate interplay between rotor and wing airflows with the ground surface~\cite{o2022neural, ajanic2020bioinspired}. 
The aerospace industry has recognized the ground effect for some time, acknowledging its potential to amplify lift forces while decreasing aerodynamic drag~\cite{rozhdestvensky2006wing, shi2019neural}. 
Despite advantages, they also pose challenges to flight stability~\cite{shi2020neural}. 
Consequently, mitigating the impacts of ground effect has been a persistent issue~\cite{shi2021neural}. 
In contrast, this paper diverges from conventional approaches by harnessing the ground effect to detect edges rather than attempting to neutralize it.
As far as we are aware, it is the first system to perform edge detection without the use of additional sensors.}

\textbf{New sensing modality.}
\majorrevise{The key insight behind this work is to translate the physical phenomenon \textit{ground effect} in flight dynamics into a fresh \textit{sensing modality} for edge detection on lightweight drones - as shown in Fig.~\ref{scenario}(b), by identifying ground effect changes, we can deduce sudden alterations in the drone's relative height above a surface or in the surface material itself, pinpointing the edge information within the scene. The process has similarities with sensing the surface by touching it using air, but only using the most basic sensing ability on lightweight drones.
However, translating this insight into a practical system still faces two challenges:}

\majorrevise{\noindent $\bullet$ \textbf{The target discrepancy between sensing and flight control complicates ground effect profiling.}
For the ground effect, the sensing modalities treat it as a \textit{friend}, aiming to detect drone flight instability through abrupt changes in sensor readings (\eg, Inertial Measurement Unit (IMU) samples) to measure it.
However, drone flight control systems view it as a \textit{foe}, striving to minimize its impact on flight dynamic stability. }
This leads to sensor (\eg, IMU) readings being extensively smoothed out after those complex proportional–integral–derivative (PID) operations~\cite{zhao2023smoothlander}, challenging the effective profiling of the ground effect.

\noindent $\bullet$ \textbf{The noisy sensing data overwhelms the vital feedback related to the ground effect.}
The dynamic measurements from inexpensive, low-power sensors on lightweight drones make it difficult to extract actual ground-effect-related fluctuations from those complex noises in raw data.
The circumstance is further complicated by the flight control system's smoothing and attenuating functions on the ground effect.

\noindent \textbf{Remark.} 
Under the premise that the flight control module treats the ground effect as a \textit{foe} and tries to negate its impact on drone stability, accurately and efficiently measuring and profiling the attenuated ground effect from noisy sensor data is crucial for edge detection.

\majorrevise{To tackle the above challenges, we design and implement \textbf{AirTouch}, the first system that treats the ground effect as a \textit{friend} and offers methods to extract related data from onboard sensors of lightweight drones, despite the flight control module's influence.}
Benefiting from AirTouch, the ground effect can be leveraged as a new sensing modality for tasks such as edge detection.
In general, AirTouch excels in the following three aspects.

\noindent \textbf{$\bullet$ On the sensory input front.}
We demonstrate that leveraging the onboard IMU readings and motor commands from flight controllers could effectively profile the ground effect. 
By examining the complex physical dynamics and drone stability control, we uncover how drone attitudes (\ie, measured by the IMU) and control signals (\ie, indicated by motor commands) interrelate and complement each other. 
\majorrevise{Their combination offers a full insight into the ground effect, even with the flight control module's adjustments.}

\majorrevise{\noindent \textbf{$\bullet$ On the algorithm front.}
We propose a ground effect-informed environmental edge detection pipeline, which comprises 
$(i)$ a fluctuation components feature extraction method and a cascaded cross-spectrum feature fusion technique to facilitate the extraction of ground-effect-related information from noisy IMU measurements and motor commands;
$(ii)$ a compact neural network (NN) designed to detect environmental edges from the extracted features;
and $(iii)$ an aerodynamics-instructed physical filter to further improve edge detection performance of neural network.}


\majorrevise{\noindent \textbf{$\bullet$ On the implementation front.}
To further boost computational efficiency and enable lightweight drones to run the proposed NN in real-time, we apply techniques such as neural unit pruning and weight quantization on the NN before onboard deployment. 
Additionally, during the NN training, we introduce a meticulously designed Disturbance Force-Informed loss function by analysis and modeling, incorporating binary cross-entropy loss, to expedite network convergence and make the network learn fine-grained bias. }

 \majorrevise{We evaluate the performance of AirTouch by conducting extensive experiments and comparing it with the baseline using a real-world testbed. Based on a lightweight drone and its onboard IMU and motors, we conducted abrupt height discontinuity edge detection and material interface transition edge detection, respectively. The results demonstrate that our system achieves a high detection accuracy with mean detection distance errors of 0.051m. Furthermore, our system surpasses the baseline performance by $86\%$ with the same available sensor information. Additionally, we compare AirTouch with vision-based edge detection methods to clarify its exclusive advantages and discuss some concerns about our system's capabilities.} Note that AirTouch is open-source on GitHub$^1$.
\renewcommand{\thefootnote}{1}
\footnotetext{https://github.com/ChenyuZhaoTHU/AirTouch}

The main contributions of this paper are as follows:
\begin{itemize}
\item We propose AirTouch, as far as we are aware, the first system that translates the traditionally negative ground effect into a new, positive sensing modality for accurate and efficient environmental edge detection.
\majorrevise{\item We demonstrate that combining IMU sampling and motor commands provides an effective sensing paradigm to characterize the ground effect under the influence of the flight control system. 
On this basis, we present a comprehensive and neural network-based pipeline aided by modeled physical knowledge for profiling, extracting, and utilizing the ground effect for sensing tasks from noisy sensory input.}
   
\majorrevise{\item We develop a prototype system and evaluate the AirTouch system through real-world data and in-field experiments on a lightweight drone by deploying our system on onboard computing chips. Extensive evaluation results show the effectiveness of our system in impressive edge detection accuracy on low-cost drones and low-energy consumption sensors and represent its exclusive advantages of capabilities by comparison with a vision-based method.}

\end{itemize}

The remainder of this paper is structured as follows: \S \ref{sec:2} presents the core intuition and prerequisites underlying the AirTouch system. \majorrevise{After introducing the system overview in \S \ref{sec:3}, we elaborate on the two main components of the system in \S \ref{sec:profiling} and \S \ref{sec:5}, respectively.} In \S 6, we introduce the implementation. In \S \ref{sec:7}, we evaluate our system. In \S \ref{sec:discussion}, we have a discussion section. In the last, we conclude this paper in \S \ref{sec:10}.

\section{Core intuitions and primers}\label{3}
\label{sec:2}

\subsection{Ground Effect and Edge Detection}
AirTouch is rooted in the features of the ground effect for a new sensing modality to detect the edge. \majorrevise{When the drone closely sweeps past a surface, its body suffers an extra airflow rebounded by the surface beneath it. The drone's flight attitude fluctuates due to the airflow caused by the ground effect. While the fluctuation from the ground effect is considered unwanted in a drone's routine flight, we leverage this physical phenomenon as exceptional feedback aids the sensing procedure.} Moreover, the variant surfaces have different attributes for the rebounded airflow, such as the reflection direction and absorption intensity. \majorrevise{Therefore, the changes in the drone's flight state under effect contain the discrepancy of variant materials, and the boundary of different levels of ground effect depicts the edges.}

\begin{figure*}[t]
\centering
\includegraphics[width=0.76\linewidth]{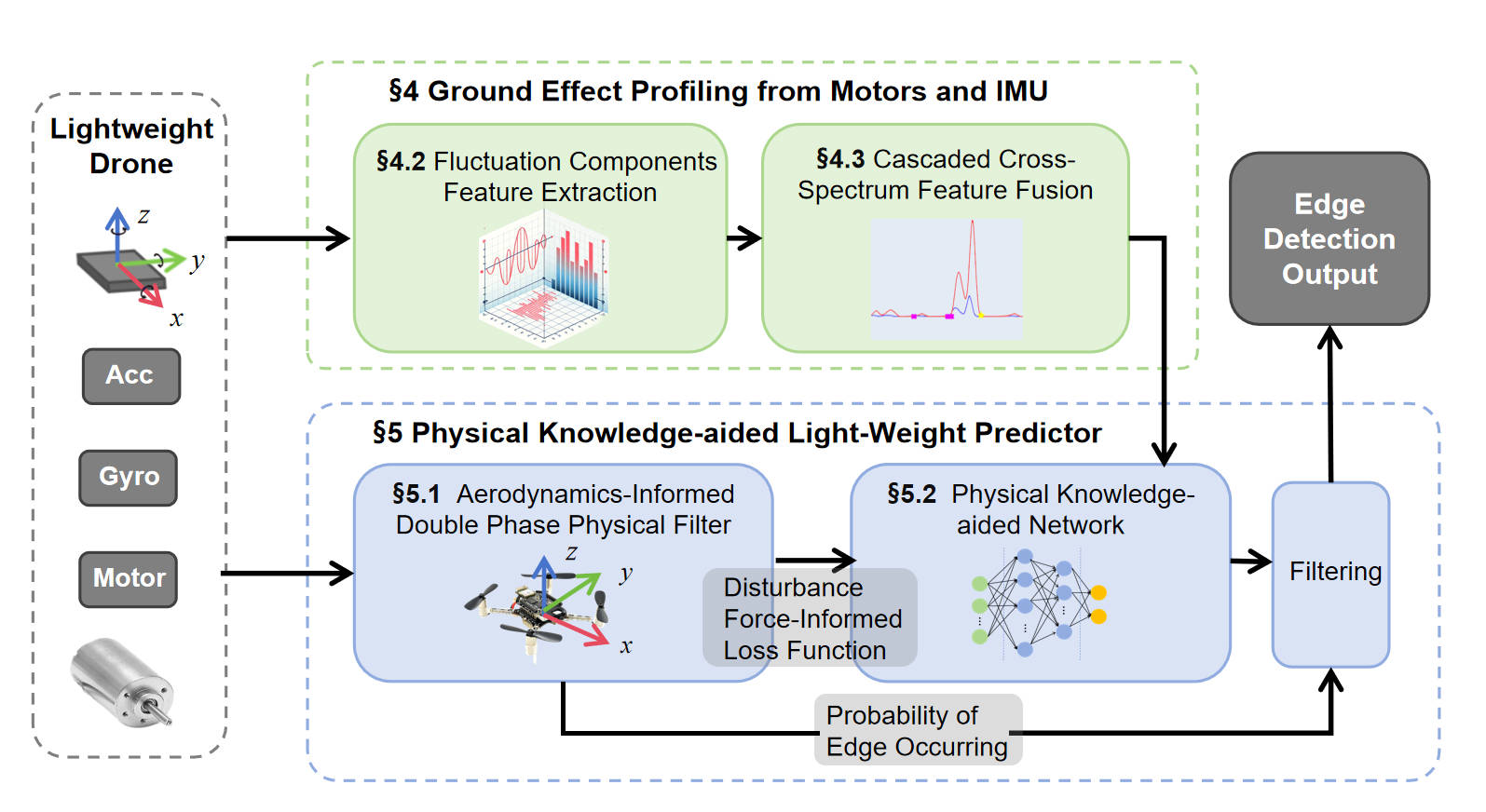}
\caption{System Overview. \majorrevise{The system is designed based on a physical-informed neural network to capture the discrepancy of the drone's state under the impact of the ground effect, to detect the scene edges. It is a new sensing modality of proprioceptive sensing, with physical phenomena.}}
\vspace{-1em}
\label{system}
\end{figure*}

\subsection{Aerodynamics of Quadrotor-like drones}
\label{Aerodynamics}
As a flight agent, the drone has states containing global position \textbf{$\mathbf{p} = [p_{x}, p_{y}, p_{z}]^\top \in{\mathbb{R}^3}$}, 
velocity $\mathbf{v}$  $\in{\mathbb{R}^3}$, body angular velocity $\boldsymbol{\omega} \in{\mathbb{R}^3}$ and attitude rotation matrix $R\in{SO(3)}$. Then, we can describe the following dynamics:
\begin{equation}
\begin{gathered}
m\mathbf{a} = m\mathbf{g} + R \mathbf{f_u} + \mathbf{f_w}, \\
\mathbf{a} = \dot{\mathbf{v}}, \mathbf{v} = \dot{\mathbf{p}}, \\
\mathbf{J}\boldsymbol{\dot{\omega}} = \mathbf{J} \boldsymbol{\omega} \times \boldsymbol{\omega} + \boldsymbol{\tau_u} + \boldsymbol{\tau_w},\\
\dot{R} = R M(\boldsymbol{\omega}), \\
\end{gathered} \label {eq:motion}
\end{equation}
where $\mathbf{a}$ is the acceleration of the drone's movement, $m$ and $\mathbf{g}$ = [0, 0, $-g$] are mass and gravity acceleration vector, respectively. $M(\cdot)$ indicates skew-symmetric mapping. $\boldsymbol{\rm{f_u}} = [0, 0, \Psi]^\top$ and $\boldsymbol{\rm{f_{w}}}$ are the forces from four rotors thrust and unknown disturbance force, respectively. \majorrevise{To simplify the formulating, if without a particular description, we assume that there is no natural wind, so the only unknown disturbance force comes from the ground effect.} $\boldsymbol{\phi} = [\Psi, {\tau_{u,x}}, {\tau_{u,y}}, {\tau_{u,z}}]^\top$ denotes the output wrench, which determines the control of quadrotors. \majorrevise{${\mathbf{u}} = [{n_1^2}, {n_2^2}, {n_3^2}, {n_4^2}]^\top$ is the actuation signal, while ${n_1}, {n_2}, {n_3}, {n_4}$ are motor rotation speeds. Accordingly, $\boldsymbol{\tau_u} = [{\tau_{u,x}}, {\tau_{u,y}}, {\tau_{u,z}}]^\top$ and $\boldsymbol{\tau_w}$ are the torques from four rotors and outside disturbance.} The thrust $\Psi$ can be derived from $\boldsymbol{\phi} = {H_0\mathbf{u}}$, with
\begin{equation}
\begin{gathered}
{H_0} = \begin{bmatrix} k_T & k_T & k_T & k_T\\ 0 & k_Tl_{r} & 0 & -k_Tl_{r} \\-k_Tl_{r} & 0 & k_Tl_{r} & 0\\ -c_Q & c_Q & -c_Q & c_Q \end{bmatrix},
\end{gathered}
\end{equation}
where $k_T$ is thrust coefficient and $l_r$ is the length of rotor arm, and $c_Q$ represents torque coefficient.
The critical factors in the drone stable flight problem are $\mathbf{f_{w}}$ and $\boldsymbol{\tau_a}$ from the unknown wind disturbance from GE. 
\majorrevise{The disturbance force $\mathbf{f_w} = [{f_{w, x}}, {f_{w,y}}, {f_{w,z}}]^\top$ and disturbance torques $\boldsymbol{\tau_a} = [{\tau_{w,x}}, {\tau_{w,y}}, {\tau_{w,z}}]^\top$ come from complicated aerodynamics interactions between quadrotors and environment, especially to the ground. In general, the larger the rotor output power or the closer the drone is to the ground, the more intensive the disturbance effect will be~\cite{alam2021survey}.}
\section{System Overview} 
\label{sec:3}

\waittoadd{From a top perspective, we design and deploy the AirTouch system to accurately detect the edges of different surfaces. The system is designed based on a physics-informed neural network to capture the discrepancy of the quadrotor's state under the impact of the ground effect, to detect the scene edges. It is a new sensing modality of proprioceptive sensing, with physical phenomena. The \textbf{core insight} is: when the drone flies closely over the boundary of two different materials, the ground effect alters due to their distinct surface properties. This allows for edge detection by monitoring the drone's posture shifts. A similar principle applies to height edge detection. The system captures implicit features from noisy raw data with minor and highly dynamic attributes. As Fig.~\ref{system} illustrates, \majorrevise{AirTouch excludes non-relevant components by extracting distinct characteristics of edges and fusing multiple features using proposed \textit{Fluctuation Component Extraction} (\S\ref{sec:4.2}) and \textit{Cascaded Cross-Spectrum Feature Fusion} (\S\ref{sec:4.3}) at first.} Then \textit{Aerodynamics-Informed Double Phase Physical Filter} (\S\ref{sec:5.1}) conducts two functionalities. The first is alleviating noisy data distraction for precision improvement with Disturbance Force-Informed Loss in \textit{Physical Knowledge-aided Network}(\S\ref {sec:5.2}). The second is filtering out false-prone edge detection from the network output. Then, the final output can be transformed into edge detection results.}

\section{Ground Effect Profiling from Motors and IMU}
\label{sec:profiling}

\majorrevise{Instead of relying on additional sensors, AirTouch utilizes a novel proprioceptive sensing method to detect edges of different surfaces by the ground effect. It is non-trivial to profile the ground effect to leverage this sensing method to distinguish the edge. Therefore, we decided to use IMU and motor signals for the detection tasks.} \textit{Combination of Two Modalities} (\S \ref{sec:4.1}) elaborates the reason by delving into the intricate physical phenomena and the flight controller working pipeline of the drone to examine its precise impact thoroughly. Then, the next challenge is how to represent the different attributes of the ground effect. A concise and interpretable representation of data captures the essential characteristics of minor GE and will enhance the detection result. Aiming at this, we proposed two sub-modules: \textit{Fluctuation Components Feature Extraction} (FC-FE) (\S \ref{sec:4.2}) and \textit{Cascaded Cross-Spectrum Feature Fusion} (CCS-FF) (\S \ref{sec:4.3}).

\begin{figure}[t]
\centering

\includegraphics[width=0.65\linewidth]{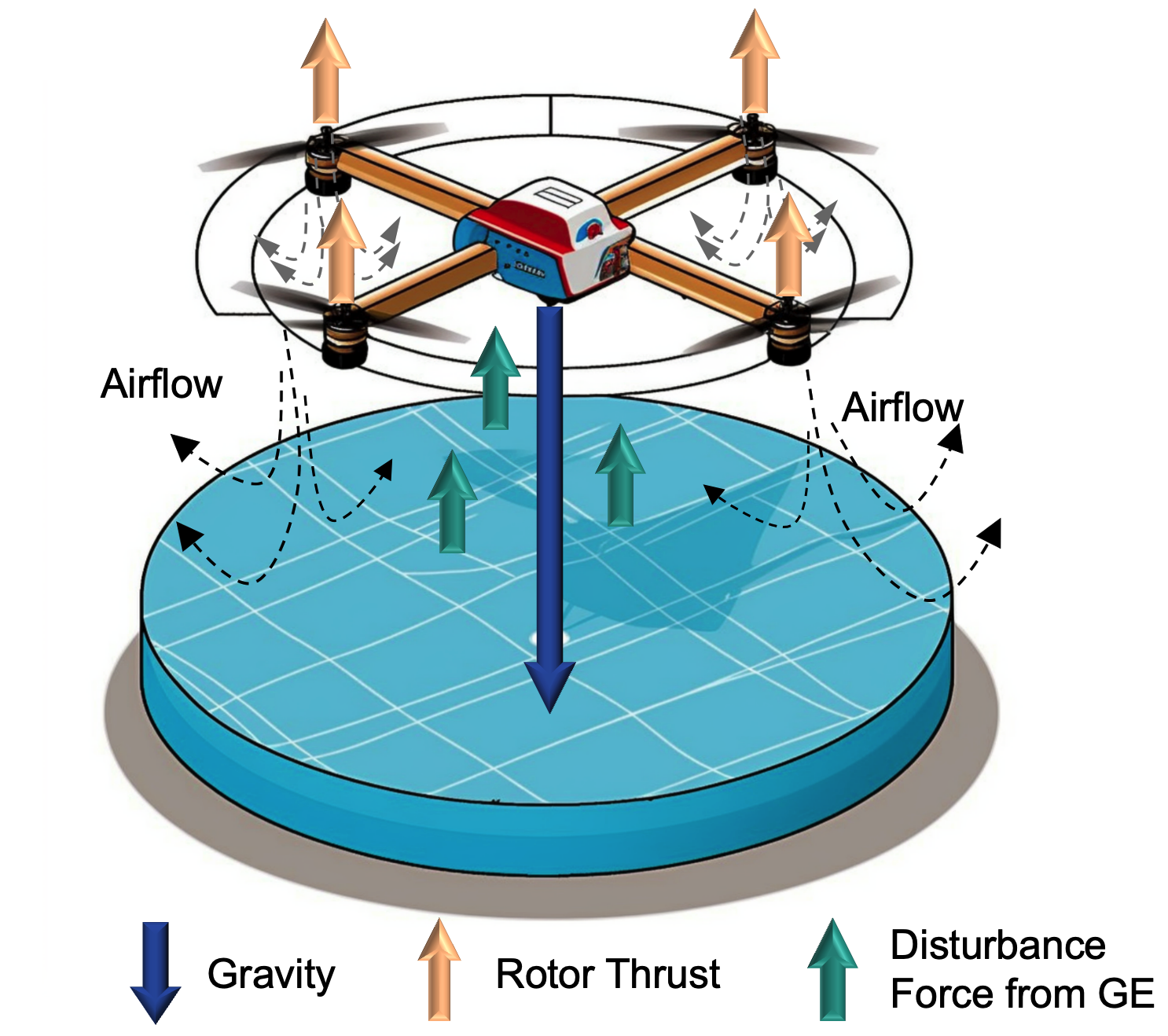}
\vspace{-0.8em}
\caption{The forces exerted on the drone under ground effect. Although forces along other axes also exist, here only the principle change components of forces, along the z-axis, are shown.}
\vspace{-0.5cm}
\label{FA}
\end{figure}

\subsection{Combination of Two Modalities}
\label{sec:4.1}
\textbf{Why both IMU and motors?}
\waittoadd{To sense the ground effect on drones, one intuitive approach is to utilize an IMU to monitor the drone's attitude changes. However, relying solely on a single modality of IMU data lacks universality from the perspective of practical engineering applications. A drone is equipped with a flight controller (FC), and one of its functions is attitude stability control. FC monitors and adjusts the drone's attitude in response to flight commands and environmental changes to maintain balance and stability. A sudden attitude change and imbalance will be captured by the IMU sensor and tentatively eliminated by adjusting the motor speeds\cite{xia2023anemoi,ebeid2018survey}. From the aspect of real-world implementation, different flight controllers (FCs) impact drone stability differently. Relying solely on IMU or motor data for ground effect detection would make the system vulnerable to variations in FC performance. For example, FCs with high precision and fast loop frequencies can mitigate attitude changes caused by the ground effect, potentially causing subtle IMU signals. Conversely, motor signals alone might not fully capture the drone's attitude changes induced by the ground effect.

From the perspective of data representation, AirTouch achieves a more comprehensive understanding of the drone's flight state data by combining IMU and motor signals. The dual data modality architecture allows the system to effectively process these complementary data sources. One data modality focuses on IMU data to capture attitude changes, while the other analyzes motor signals to understand the flight controller's adjustments. This design ensures that the system can reliably detect ground effect changes across different FCs, enhancing its generalizability and robustness.}

\majorrevise{Thus, motor signals and IMU's attitude information \textbf{are coupled and complementary to each other}, together contributing to represent the whole effect of the ground effect. To ensure that our detection method based on ground effect supports drones with different flight controllers, we profile the ground effect using both IMU and motor modalities from both a physical force analysis perspective and an input-data characteristic perspective.}

\begin{figure*}[ht]
\centering

\includegraphics[width=\linewidth]{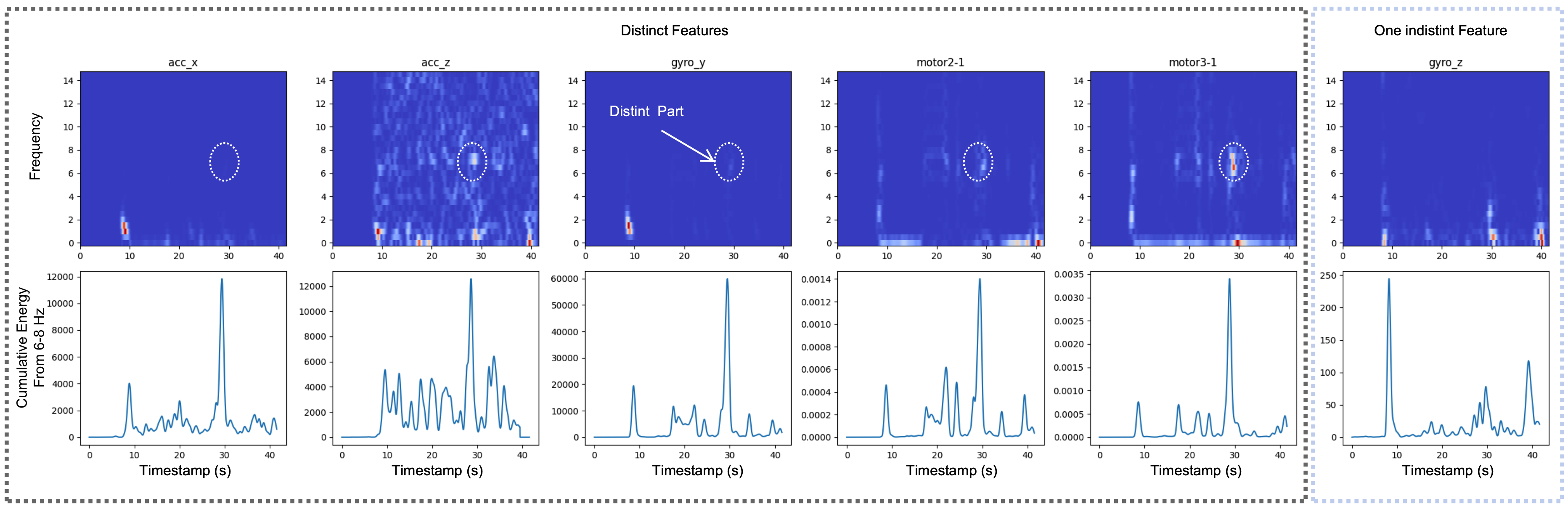}
\vspace{-0.8em}
\caption{The left five plots show the distinct features, and the right one is an example of an indistinct feature. From the heatmaps of the power spectrum, the target periods in the left features coherently have distinct power increases between 6-8 Hz, while it may not be distinct or stable among other features.}
\label{features}
\end{figure*}

From a physical force analysis perspective, we derive the disturbance force $\mathbf{f_w}$ as the direct impact measurement of the ground effect. \majorrevise{As is shown in Fig.~\ref{FA}, we exclude the gravity and thrust force from net force to derive the disturbance force using Newton's second law:}
\begin{equation}
\begin{gathered}
\mathbf{f_w} = m\mathbf{a} - m\mathbf{g} - R \mathbf{f_u}, \\
{\Psi} = H_0 W(\mathbf{m}), \ \dot{R} = R M(\boldsymbol{\omega}), \\
M(\boldsymbol{\omega}) = \begin{bmatrix} 0 & -gyro_z & gyro_y \\ gyro_z & 0 & -gyro_x \\ -gyro_y & gyro_x & 0 \end{bmatrix},
\end{gathered} \label {eq:fw}
\end{equation}
where ${W(\cdot)}$ is the transform mapping from motors' PWM signals $\mathbf{m} = [m_1, m_2, m_3, m_4]^\top$ to actuation signal $\mathbf{u}$. Apart from system parameters, data sources needed in the above derivation depend on $\mathbf{a}$ and $\boldsymbol{\omega}$ measured by IMU and $\mathbf{m}$ sent to motors.

Additionally, considering the data characteristic, two modalities contribute to data-driven edge detection. From the feature extraction results introduced later, we discover that the influence of the drone attitude and motor signals has dynamicity, which means the intensity of the ground effect varies across different detection trials. For this concern, redundancy in data sources could reduce the detection failure rate. At the same time, IMU and motors are delayed and feedback to each other. \majorrevise{Therefore, to enable a comprehensive understanding of the drone's attitude and empower edge detection performance, our system aims to contain the thorough ground effect's features, by leveraging both motor signals and IMU signals as the source data.}

\textbf{What is data dimension?} The attitude change induced by GE is captured by the flight data of the drone, including the 3-axis acceleration rate $\mathbf{a} = [acc_{x}, acc_{y}, acc_{z}]^\top$ from the acceleration measurement, 3-axis angular acceleration rate $\boldsymbol{\omega} = [gyro_{x}, gyro_{y}, gyro_{z}]^\top$ from the gyroscope, and four motors' PWM signals $\mathbf{m} = [m_1, m_2, m_3, m_4]^\top$ calculated from command signals. Here, $\mathbf{m}$ is positively proportional to the motors' rotation speeds $[n_1, n_2, n_3, n_4]^\top$. \majorrevise{Fig.~\ref{raw_data} illustrates a set of the raw data collected during a drone's flight.}

\subsection{Fluctuation Components Feature Extraction (FC-FE)} 
\label{sec:4.2}
\majorrevise{We propose the FC-FE algorithm, a frequency spectrum analysis-based technique to extract features of high sensitivity and accuracy.} It is based on the fluctuation of the attitude and motor signals. The temporal data source is $\mathbf{d} = [acc_{x}, acc_{y}, acc_{z}, gyro_{x}, gyro_{y}, gyro_{z}, m_{2-1}, m_{3-1}, m_{4-1}]^\top$. For normalization, $m_{2-1}$, $m_{3-1}$, and $m_{4-1}$ are the differences between other three motors and $m_1$. Note that, for the convenience of explanation, we \textbf{take the height edge detection as the example case} in \S \ref{sec:profiling} and \S \ref{sec:5}. We first examine the heat maps of power amplitude versus time to explore the useful frequency components over time in all 9 feature categories, which are transformed by Short-Time Fourier Transform (STFT)~\cite{wang2021distracted}. We analyze the sum power amplitude component $P$ within a distinct frequency band $F$. $P$ satisfies the fact that it is distinguishing at the edge. \majorrevise{As an example illustrated in the left five heat maps in Fig.~\ref{features}, the power amplitude from 6 to 8 Hz is always distinctly stronger when the drone sweeps past the edge in most data categories.} The consistency lies in the rigid contact of IMU sensors and motors during the drone fluctuation. This particular power cluster reflects the attitude oscillation of the drone under the ground effect, which is determined by many factors, such as the drone’s weight, size, propeller size, performance, and bias of the flight controller, etc. Theoretically, such
frequency range differs from different drone types and our tests with larger drones coincide with the observation. \majorrevise{Considering generality and universality, the frequency range can be easily collected and configured on the system.}

We calculate the cumulative power amplitude within $F$ at each timestamp and take the results as features to distinguish the edge. \majorrevise{Correspondingly, the smaller the temporal and frequency resolutions are, the better performance of real-time capability and detection precision the system may have.} Note that the resolutions of frequency and temporality correlate to the window size and overlap size of STFT.

\subsection{Cascaded Cross-Spectrum Feature Fusion (CCS-FF)} \label{sec:4.3}
\majorrevise{We design Cascaded Cross-Spectrum to fuse multiple features of each sensor into one synthetic feature, and then three independent synthetic features for acceleration rate, angular acceleration rate, and motor signal will be generated.} Inspired by the cross-spectrum, which is a method to assess the correlation between the components of two signals at the same frequency, CCS-FF extracts the correlation within one sensor's different dimensions to represent a comprehensive characteristic. 

Before introducing CCS-FF, we first introduce the cross-spectrum~\cite{nelson2001cross}. If there is a strong correlation between the two signals at a particular frequency, the amplitude of the cross-spectrum at the corresponding frequency will be large. $G_{xy}(f)$ is the cross-spectrum of two signals $x(t)$ and $y(t)$, whose frequency domain represents $X(f)$ and $Y(f)$. The magnitude component $|G_{xy}(f)|$ quantifies the degree of correlation between the two signals at the frequency $f$. $G_{xy}(f)$ is defined as the product of $X(f)$ and the conjugate of $Y(f)$, as shown in the following Eq.\ref{eq:cross-spectrum},
\begin{equation}
\begin{gathered}
G_{xy}(f) = X(f)Y^*(f).
\end{gathered} \label {eq:cross-spectrum}
\end{equation}

Meanwhile, as a signal's power spectral density, a fundamental concept used to characterize the power distribution of a signal across various frequencies in the frequency domain can be expressed as Eq.\ref{eq:power-spectrum}; thus, the relationship between $|G_{xy}(f)|$, $G_{xx}(f)$ and $G_{yy}(f)$ satisfies the relationship as defined in Eq.\ref{eq:abs-spectrum}.
\begin{equation}
\begin{gathered}
G_{xx}(f) = X(f)X^*(f).
\end{gathered} \label {eq:power-spectrum}
\end{equation}
\begin{equation}
\begin{gathered}
|G_{xy}(f)|^2= |G_{xx}(f)||G_{yy}(f)|.
\end{gathered} \label {eq:abs-spectrum}
\end{equation}

To this end, the correlation between signals can be calculated using the product of their power spectral densities. Extending this concept to more than two signals involves cascading their cross-spectrum. For $n$ signals with frequency domain representations $X_1(f)$, $X_2(f)$, ..., $X_n(f)$, we define the cascaded cross-spectrum $P_{ccs}(f)$ and use its magnitude to measure the degree of correlation between the $n$ signals, which can be represented as if $n$ is an even number, 
\begin{equation}
\begin{aligned}
|P_{ccs}(f)| &= |X_{1}(f)X^*_{2}(f)X_{3}(f)...X^*_{n}(f)|
\\&= \sqrt{|G_{x_1x_1}(f)||G_{x_2x_2}(f)|...|G_{x_nx_n}(f)|}.
\end{aligned} 
\label {eq:mul-cross-spectrum}
\end{equation}

Note that if $n$ is an odd number, the last term of the first row should be $X_{n}(f)$. Based on the aforementioned derivation, the power spectral densities of the extracted features are multiplied to serve as the synthetic fusion feature providing correlation information to represent the comprehensive characteristic. To be specific, 9-dimensional temporal data $\mathbf{d}$ has been transformed into 3-dimensional synthetic NN input data $\mathbf{c} = [\mathbf{a_s}, \boldsymbol{\omega_s}, \mathbf{m_s}]^\top$.


\section{Physical Knowledge-aided Light-Weight Predictor} 
\label{sec:5}
To facilitate the drone's real-time and accurate acquisition of edge information, we employed a lightweight neural network (NN) model for edge prediction. As discussed in the previous section, the coupling of motor and IMU data challenges the precise capture of the ground effect. Compared to traditional methods, NN can better capture the spatio-temporal relationship between drone sensor fluctuations and ground effect. By modeling the ground effect, extracting and fusing features in \S \ref{sec:profiling}, we obtained the input for the NN (\S \ref{sec:5.2}). Moreover, we utilized the \textit{Aerodynamics-Informed Double Phase Physical Filter} proposed in \S \ref{sec:5.1} to guide the network in learning data more efficiently. Given the limited computing and storage resources of the drone, we also conducted lightweight processing on the model in \S \ref{sec:5.3}. 

\majorrevise{\subsection{Aerodynamics-Informed Double Phase Physical Filter}
\label{sec:5.1}
Aerodynamics-Informed Double Phase Physical Filter consists of two functionalities in two phases: training phase and output phase of the neural network. One functionality is that it provides a fine-grained bias serving as a physical knowledge-based loss item in the NN training phase for alleviating noisy data distraction, to improve detection precision. The other one is that it minimizes the occurrence probability of false detection caused by edge-irrelevant disturbances in the final output phase. The core intuition is leveraging the inherent disturbance force exerted on the drone, hidden in the distinction between ground effects separated by the edge. In both phases, \textit{Fast Responded Constant False Alarm Rate Algorithm} (FR-CFAR), serves as an adaptive threshold selector for the enlightenment of the extra force change could be detected.

Inspired by the aerodynamics analysis introduced in \S \ref{Aerodynamics} and Eq. \ref{eq:fw}, we derive and then leverage the unknown disturbance force $\boldsymbol{\rm{f_{w}}}$ exerted on the drone body to get a glimpse of the height edge detection process. The $\mathbf{f_w} = [{f_{w, x}}, {f_{w,y}}, {f_{w,z}}]^\top$ is induced by the extra disturbance of the rebound airflow solely and here we only focus on the upward force $f_{w,z}$ because the distinct component of force change is along $z$ axis in height edge detection. It is similar in cases of material edge detection, where the distinct change of force may occur along other axes. Overall, as the core physical information, $\mathbf{f_w}$ can be used to denote the physical characteristic of the detected surface.

As is shown in Fig.~\ref{FaZ}, the disturbance force in the $z$ axis is plotted over the processes of idle, takeoff, hover, and traveling across the platform. In the beginning, the $f_{w,z}$ is equal to the supporting force from the ground, whose value is equal to the gravity of the drone. Then, when the drone is hovering or flying in the air, this value of the extra force is around zero. Only when the drone is sweeping over the platform, $f_{w,z}$ have significant increasing peaks roughly from 29.37 to 32.81 seconds in the plot, roughly consistent with the ground truth. Although it is difficult to capture a clear pattern with relevant constant values representing the detected surface, it is obvious that the disturbance force reflects the surface information and the significant amplitude change represents the boundary of surfaces with different attributions.

\begin{figure}[t]
\centering
\vspace{-0.8em}
\includegraphics[width=0.98\linewidth]{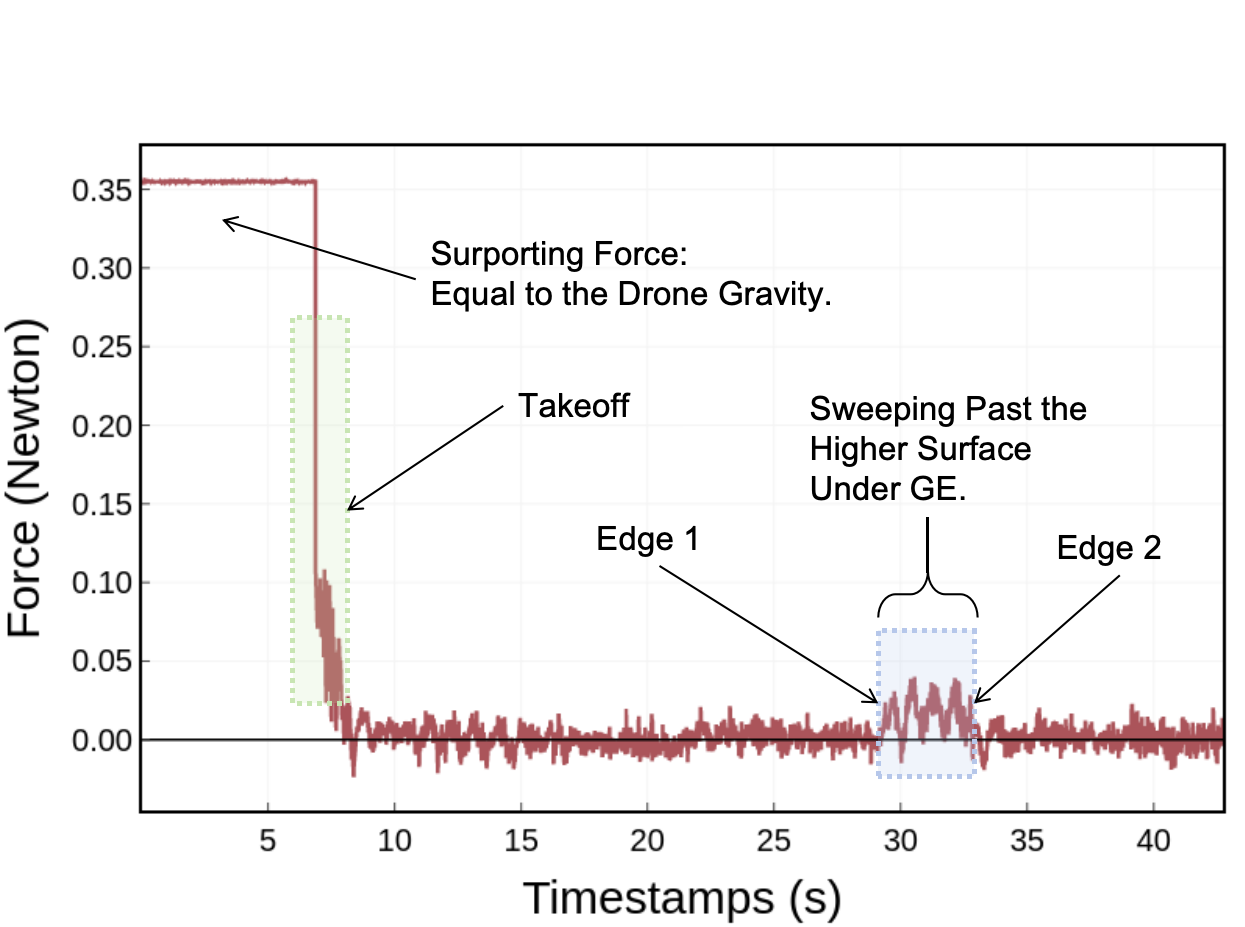}
\vspace{-0.3em}
\caption{Disturbance force along the z-axis in height edge detection in Aerodynamics-Informed Double Phase Physical Filter. The magnitude of extra upward lift $f_{w,z}$ is the data source of the physical filter. The convex part of the force represents that the drone suffers an extra upward lift due to the ground effect.}
\vspace{-1.5em}
\label{FaZ}
\end{figure}

To apply physical embedding for the neural network and exclude the false edge detection caused by data noise or non-relevant flight operation, a strategy is necessary to capture the mutation of the disturbance. However, limited by the environmental parameter accuracy and sensor precision, the $f_{w}$ measurement may have offset and noise. This makes it impossible to notice the mutation with a simple constant threshold. Therefore, we propose the Fast-Respond Constant False Alarm Rate method to adjust the threshold and recognize the force mutation. 

\textbf{Fast-Respond Constant False Alarm Rate (FR-CFAR).} 
As is illustrated in Fig.~\ref{FR-CFAR}, an FR-CFAR algorithm is proposed based on procedures of the Cell-Averaging Constant False Alarm Rate (CA-CFAR) to select the target points, which provides a constant false detection probability~\cite{jalil2016analysis}. FR-CFAR detects target signals in background noise while ensuring a constant probability of false positives, and also has instant detection capability. The difference between a normal CA-CFAR and our FR-CFAR is the omission of the lagging window, which requires queue time for the following signal that has not been generated yet. Therefore, without waiting for the lagging window data, FR-CFAR can respond and output the detection result instantly.

\begin{figure}[t]
\centering
\includegraphics[width=0.85\linewidth]{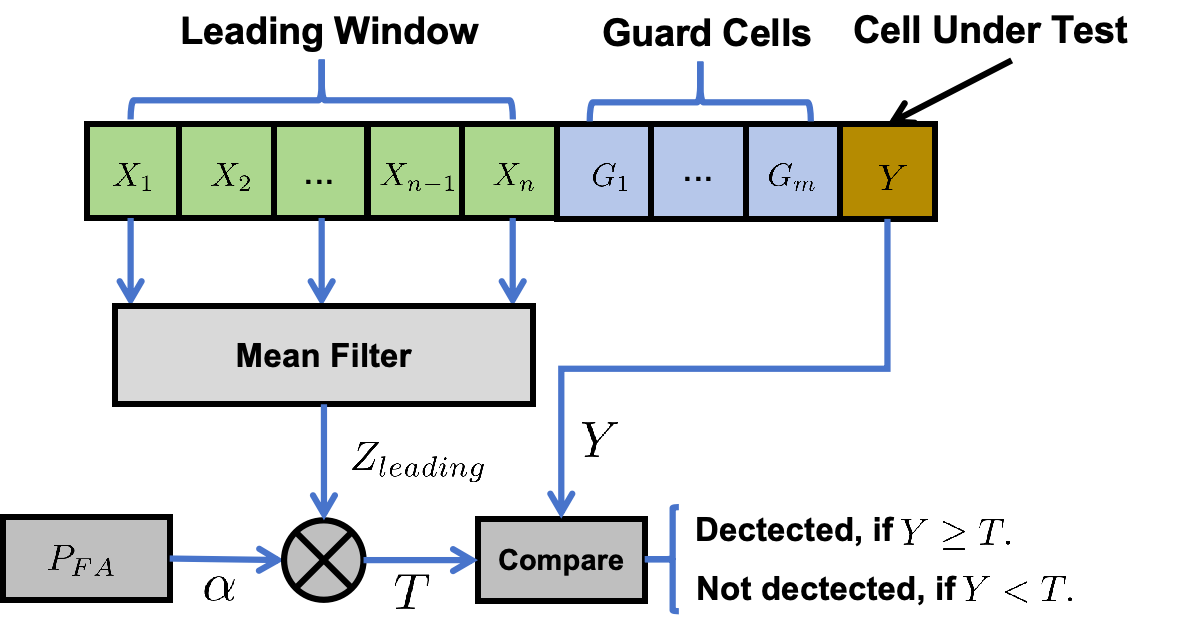}
\caption{The main procedure of Fast Responded Constant False Alarm Rate (FR-CFAR) algorithm.}
\vspace{-1.5em}
\label{FR-CFAR}
\end{figure}

The key scheme is that the algorithm proposes a threshold level calculated by estimating the noise floor level around the judged signal sample, cell under test (CUT). If the signal sample's magnitude exceeds the threshold, the CUT $Y$ is considered the target. A group of cells surrounding the Cell Under Test (CUT) is selected to determine this, and the average power level is calculated. $[G_1, ..., G_m]$, the cells directly adjacent to the CUT, known as "guard cells", are typically excluded from this calculation to prevent interference from the CUT itself, thus preventing a more clean estimate from adjacent interference. The detection threshold $T$ satisfies the following relationships:
\begin{equation}
\begin{gathered}
Z_{leading} = \frac{1}{N} \sum_{i=1}^{N} X_i,\\
T = \alpha \cdot Z_{leading},\\
\alpha = N*(P_{FA}^{-\frac{1}{N}}-1),
\end{gathered} \label {eq:CFAR}
\end{equation}
where $Z_{leading}$ is the mean signal value of the $N$-cell leading window to estimate the noise level, $X_i$ denotes the signal received in the $i$ th cell in the leading window, and threshold factor $\alpha$ is adjusted to maintain the false alarm rate at a predetermined probability $P_{FA}$~\cite{farina1986review}. Once the CUT $Y$ exceeds the dynamic detection threshold $T$, the disturbance force alerts a high probability of edge occurring. Correspondingly, this filter improves detection precision for NN training in the first physical embedding phase by alleviating noisy data distraction. For the second phase, the final NN predictor detection without alerts from disturbance force will be filtered and ignored.

One important thing that needs to be further clarified is the necessity of the combination of physical knowledge (disturbance force) and data-driven method (neural network). On the one hand, although the disturbance force directly discloses the physical characteristics of the ground effect and edge information, it has weaknesses hinder the accuracy of the exact edge detection. Due to the noisy sensing data from the IMU and motor signals, it is difficult to capture the accurate moment of the edge occurrence, limited by the performance of low-cost sensors. The significant oscillation in amplitude always begins slightly lagging behind the exact moment of the edge occurrence (Fig.~\ref{raw_data}), which causes larger detection errors. This delay may come from the topology and weight distribution of the drone and the interaction conditions with the edge (i.e., height, angle, velocity). Therefore, the neural network can provide powerful nonlinear fitting capability to decrease error in detection tasks, as a complement to ensure the physical knowledge efficiency. On the other hand, purely relying on data-driven the neural network may require a large amount of data for training and suffer from unstable output. Therefore, we use physical knowledge to constrain the output of the neural network, and optimize training process to expedite network convergence and make it learn fine-grained bias.}

\subsection{Physical Knowledge-aided Network} 
\label{sec:5.2}

The input of our model consists of three-dimensional fused features denoted as $\mathbf{c} = [\mathbf{a_s}, \boldsymbol{\omega_s}, \mathbf{m_s}]^\top$ within a specific time window, resulting in an input size equal to three times the window size. The model is designed for binary classification, where the detection of edges is accomplished by identifying transitions between the two encoded categories.

\waittoadd{To address the specific requirements of edge detection and energy conservation in our task, we employ a standard NN architecture. The input is reshaped from (100, 3) to (100, 3, 1) to prepare for 2D convolution. The first Conv2D layer has 16 filters of size (3, 1) with ReLU activation, followed by a MaxPooling2D layer with pool size (2, 1). The second Conv2D layer has 32 filters of the same size and activation function, followed by another MaxPooling2D layer. The Flatten layer then converts the output to 1D. Two Dense layers follow: one with 16 neurons and ReLU activation, and an output layer with 2 neurons and sigmoid activation for classification. The output layer of the model consists of two units with softmax activation, enabling the model to generate class probabilities for each category.}


\textbf{Disturbance Force-Informed (DF) Loss Function. }The incorporation of a physics-informed loss function aims to enhance the model's comprehension of ground effect dynamics. When a drone encounters an edge, a previously unknown disturbance force undergoes an abrupt change, serving as a distinctive indicator for edge detection. By integrating this force in conjunction with a threshold, our physics-embedded methodology effectively captures relevant features associated with ground effects. This approach improves the precision of edge detection and yields valuable insights into the flight environment of the drone.

For instance, considering a drone traversing the edge of an abrupt height change, a state variable switches between 1 and 0 depending on the surface. The edge marks the transition between these states, with distinct disturbance forces experienced on each surface. The disturbance force-informed loss function employed is computed as follows:

\begin{equation}
     \mathcal{L}_F= 
     \begin{cases}e^{\left| \hat{y}-1\right|}-1, & ||f_w|| \geqslant T. 
     \\ e^{\left| \hat{y}\right|}-1, & \text {others }. \end{cases},
\end{equation}
where $\mathcal{L}_F$ denotes the physics-informed loss function, which is designed to measure the model's prediction error for disturbance forces. The $ \hat{y}$ represents the predicted state, and the $T$ is the threshold used to determine if the disturbance force $f_w$ exceeds a critical value. The disturbance force $f_w$ and $T$ can be calculated in \S\ref{sec:5.1}.

For this binary classification problem, we utilize the binary cross-entropy as our fundamental loss function:
\begin{equation}
    \mathcal{L}_S =-y \log \hat{y}-(1-y) \log (1-\hat{y}),
\end{equation}
where $y$ is the sample's actual label (0 or 1), and $\hat{y}$ is the predicted probability value output by the model.

Considering the two loss functions, the overall loss function for edge detection is:
\begin{equation}
     \mathcal{L} = \mathcal{L}_S  + \lambda \cdot \mathcal{L}_F,
\end{equation}
where $\lambda$ is the coefficient that weighs the contribution of the disturbance force-informed loss function and can be adjusted easily in different applications.

\majorrevise{
\subsection{Pruning and Quantization:}
\label{sec:5.3}
For the lightweight design of our model, we employed pruning and quantization to reduce complexity, enhance efficiency, and improve performance, making it well-suited for resource-constrained platforms.

Weight pruning is a technique aimed at reducing the computational complexity of neural networks by eliminating redundant connections. 
In our pruning procedure, weights are systematically pruned based on an adjustable threshold. Specifically, we first sort the absolute values of all weights in ascending order to obtain a vector $\Lambda = \{W_1, \ldots, W_N\}$, where $N$ denotes the number of convolutional layers and $W_n$ denotes the kernel weights used by the $n$th layer, $n = 1, \ldots, N$. We then calculated the threshold, denoted as $\theta$. This threshold is used to prune weights based on their magnitude. The threshold is dynamically adjusted during training using a pruning schedule, represented by the formula:
\begin{equation}
    \theta = s_i + (s_f - s_i) \times \left(1 - \frac{t_i}{t_e}\right)^{p},
\end{equation}
where $s_i$ is the initial sparsity level, $s_f$ is the final sparsity level, $t_i$ is the current training step, $t_e$ is the total number of training steps, and $p$ is the power of the polynomial. Each element of $\Lambda$ is compared to this threshold, and weights below the threshold are pruned according to the formula:
\begin{equation}
    [W_n]_i = \begin{cases} 
    [W_n]_i, & \text{if } |[W_n]_i| \geq \theta \\
    0, & \text{otherwise},
    \end{cases}
\end{equation}
where $[W_{n}]_{i}$ denotes the $i$th element of $W_{n}$ in a certain order for $n = 1, \ldots, N $ and $i = 1, \ldots, L_{n}$ with $L_{n}$ representing the total number of elements of $W_{n}$. 
This systematic approach, in which weights are either retained or set to zero based on their magnitude compared to the threshold, enables effective weight pruning in convolutional neural networks, leading to model compression. 

Pruning aims to remove unimportant components, and Quantization involves converting float32 weights or activations to lower-precision floating-point numbers or integers. While pruning removes many weights, the ones that remain are still represented as 32-bit floating-point numbers. This can lead to excessive computational requirements and storage space usage and can be resolved by quantization.

Quantization is the establishment of a mapping relationship between floating-point data and fixed-point data. We denote floating-point real numbers as $ r$ and quantized fixed-point integers as $ q $. The conversion formula between floating-point and integer is given by:
\begin{equation}
    \begin{aligned}
    r &= S(q-Z) \\
    q &= \text{round}(r / S) + Z,
    \end{aligned}
\end{equation}
where $ S $ is the scale factor representing the proportionality between real and integer values, and $ Z $ is the zero point representing the integer equivalent of 0 in real numbers. The $round$ function rounds the result of $r \times S$ to the nearest integer. These are calculated as follows:
\begin{equation}
    \begin{aligned} 
     S &= \frac{\text{max}(r) - \text{min}(r)}{\text{max}(q) - \text{min}(q)}, \\
     Z &= \text{round}(-\text{min}(r)/S), 
    \end{aligned}
\end{equation}
where the $min$ and $max$ functions return the minimum and maximum values respectively. After weight pruning and quantization, the lightweight CNN model is obtained with a model compression
ratio:
\begin{equation}
    CR = \frac{S_o}{S_c} = \frac{32}{(1 - s_f) \times {b}}
\end{equation}
where the original model size is $S_o$, $S_c$ is the size of the model after pruning and quantization, the pruning sparsity level is $s_f$, and the number of quantization bits is 
$b$.
}
\section{Implementation}
\label{Implementation}
\begin{figure}[t]
\centering
\includegraphics[width=0.99\linewidth]{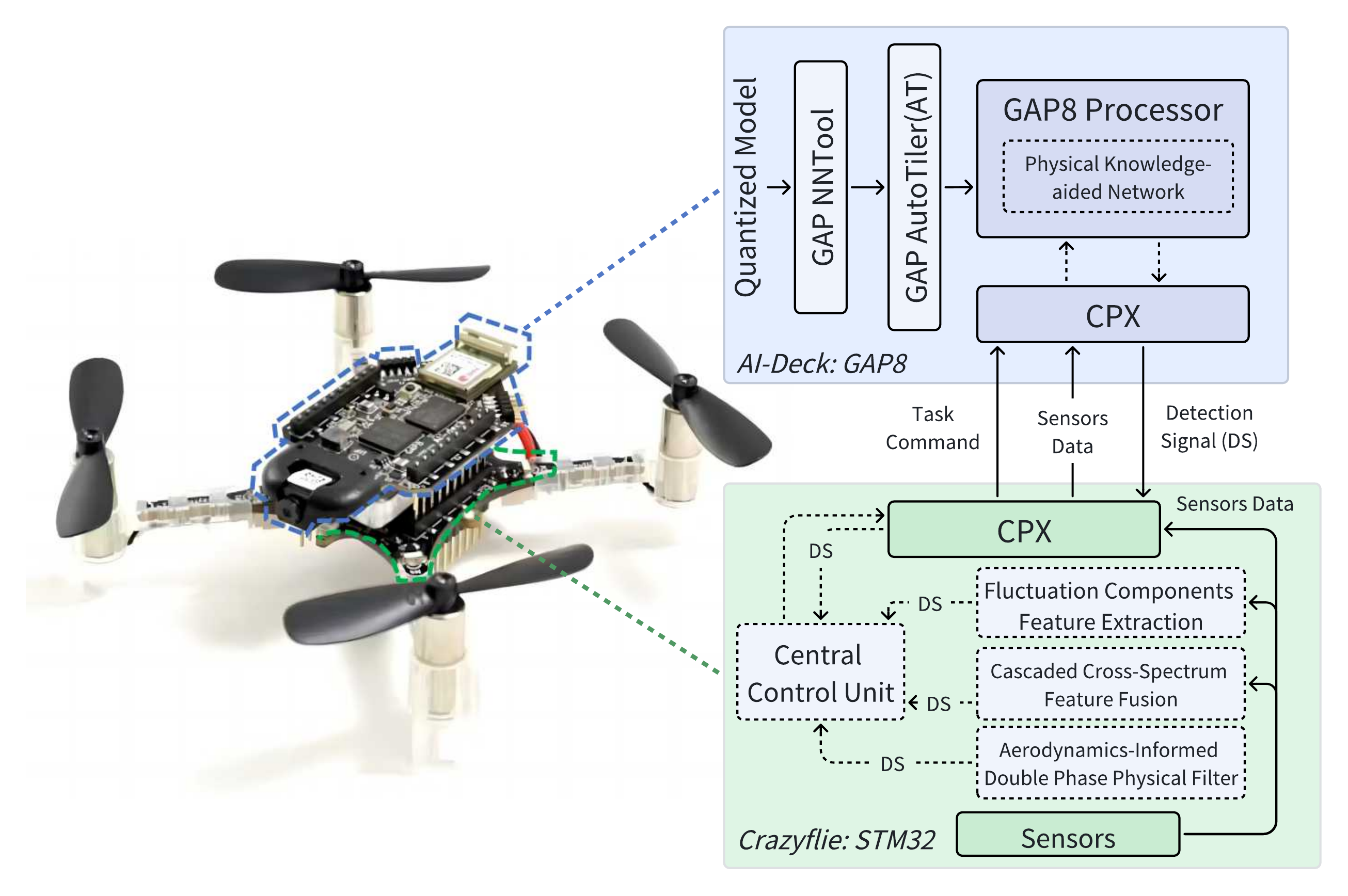}
\caption{Implementation framework and modules on the devices.}
\vspace{-2em}
\label{ImplementationFig}
\end{figure}

To experimentally validate our system for precise edge detection, we selected the Crazyflie 2.1 nano-quadrotor, including an STM32F405 MCU
and a 250mAh LiPo battery. It weighs 35.6 grams and offers 22.65 GOPS of processing power and 512KB of RAM. Additionally, our test drone is equipped with an AI deck, a Multi-ranger deck, and a Flow deck. The AI deck, equipped with a GAP8 microcontroller, enhances the drone's memory and computational capabilities, which is crucial for implementing our NN model. 
The Multi-ranger deck detects objects around the Crazyflie, while the Flow deck tracks the drone's movements.

We implemented the \textit{Fluctuation Components Feature
Extraction} (\S\ref{sec:4.2}), \textit{Cascaded Cross-Spectrum Feature
Fusion} (\S\ref{sec:4.3}) and \textit{Aerodynamics-Informed Double Phase Physical Filter} (\S\ref{sec:5.1}) on the STM32 and then fed the fusion data into our \textit{Physical Knowledge-aided Network} (\S\ref{sec:5.2}) on the AI deck, as is shown in Fig.~\ref{ImplementationFig}. The AI deck communicates with the STM32 on the Crazyflie to obtain the fusion data for the actual model inference implementation. 

\section{Evaluation}
\label{sec:7}
To this end, we implement a prototype of AirTouch and perform experiments with different scenes and parameters. We first introduce the experimental setup and evaluate the overall performance of our system. Next, we present two case studies demonstrating our system's effectiveness and accuracy in detecting different types of edges. Additionally, we explore the system's resilience and evaluate specific components' contributions. \majorrevise{Then, we measured the model's weight size, inference speed, and energy consumption to demonstrate the efficiency of our system. Finally, we compare our proposed system with vision-based methods to clarify its exclusive advantages.}

\begin{figure}[t]
\vspace{-0.5em}
\centering
\includegraphics[width=0.95\linewidth]
{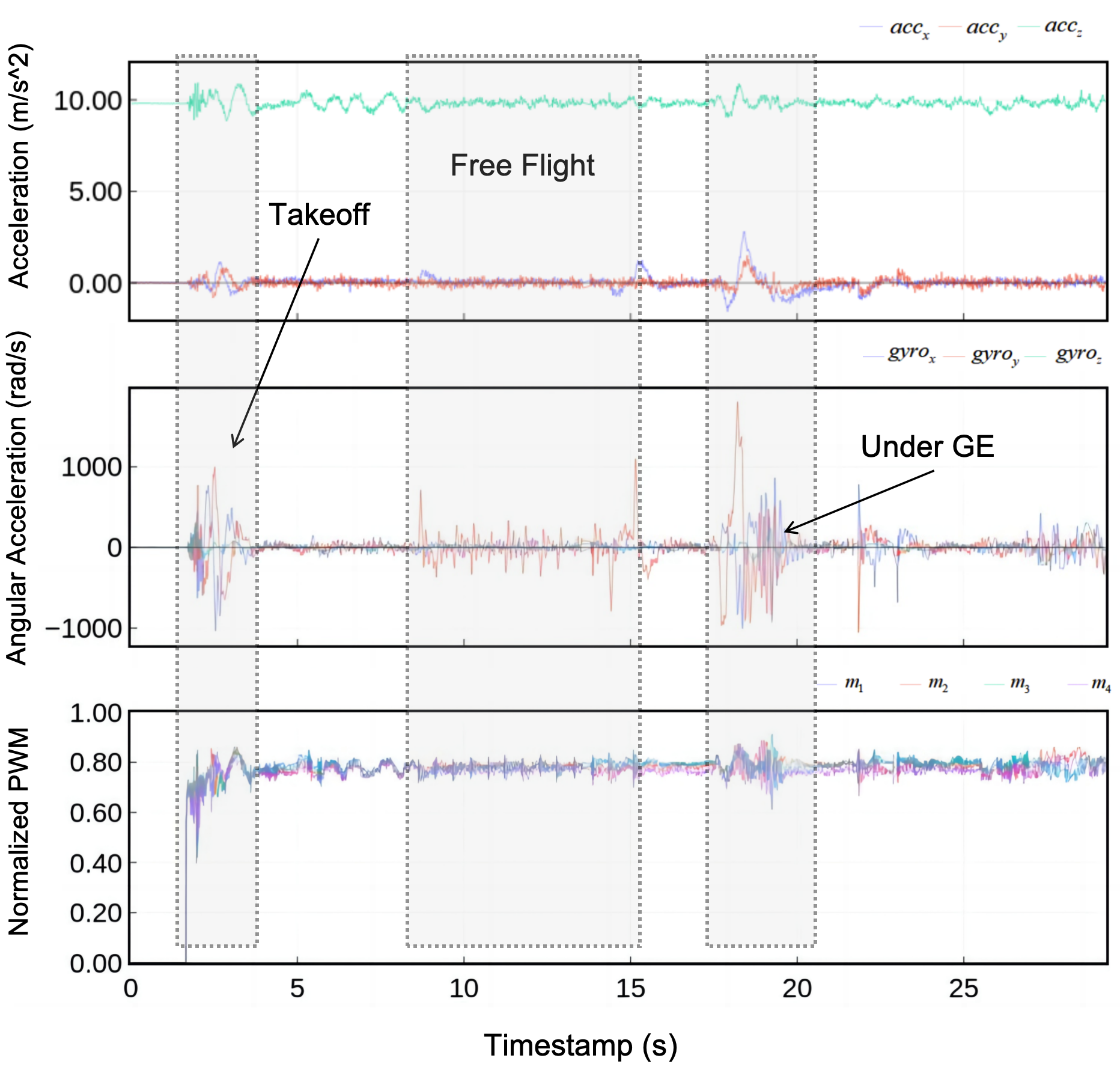}
\vspace{-1em}
\caption{The raw data used in the system. The data contains the acceleration rates, angular acceleration rates, and the motor PWM signals from takeoff, normal flight, to sweeping past the platform surface.}
\vspace{-1.5em}
\label{raw_data}
\end{figure}

\begin{figure*}[htbp]
    \centering
    \vspace{-0.5cm}
    \subfloat[Errors of Abrupt Height Discontinuities]{%
        \includegraphics[width=0.325\linewidth]{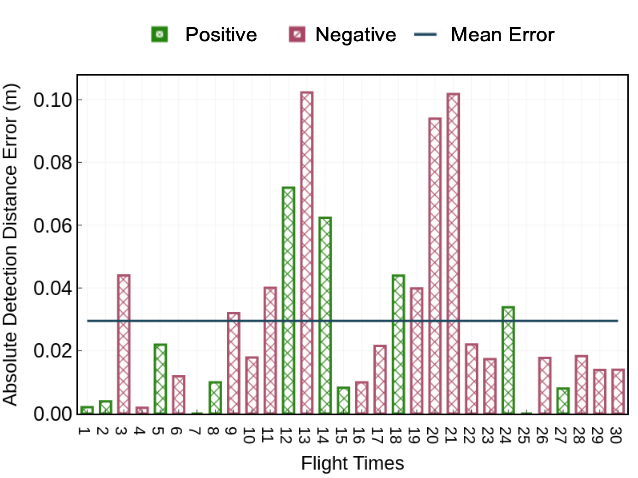}%
        \label{Performance:1a}
    }\hfill
    \subfloat[Errors of Material Interface Transitions]{%
        \includegraphics[width=0.325\linewidth]{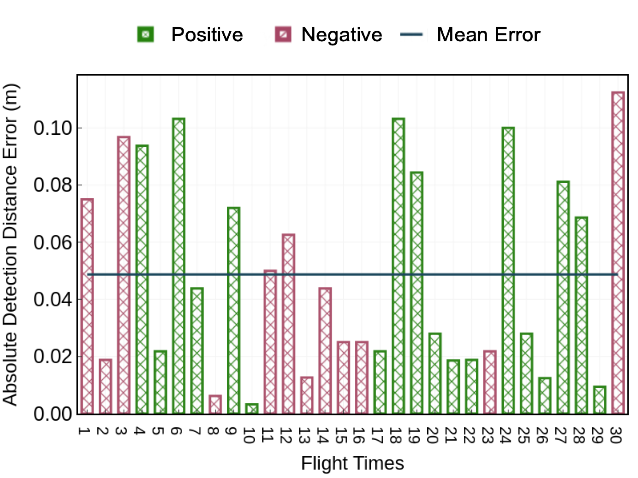}%
        \label{Performance:1b}
    }\hfill
    \subfloat[Errors of Our System and Baseline]{%
        \includegraphics[width=0.325\linewidth]{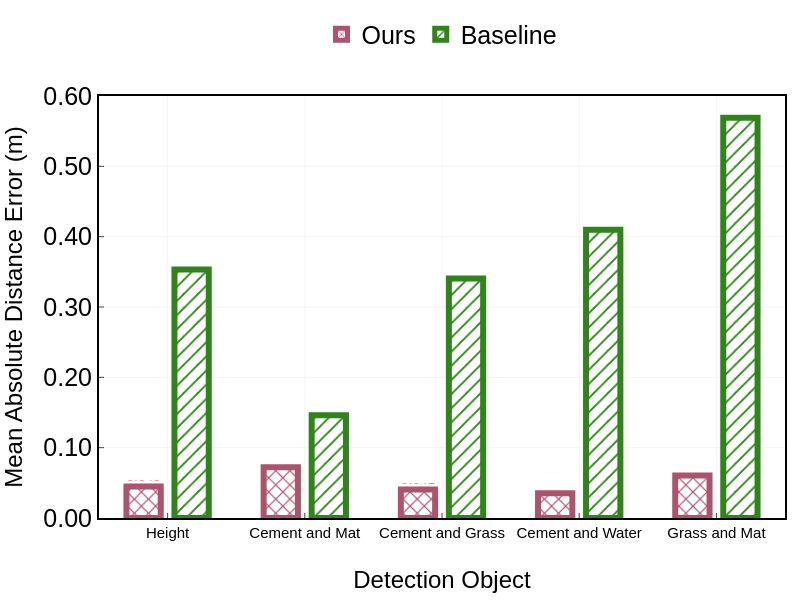}%
        \label{Performance:1c}
    }

    \vspace{-0.2cm}
    \subfloat[CDF of Different Heights]{%
        \includegraphics[width=0.325\linewidth]{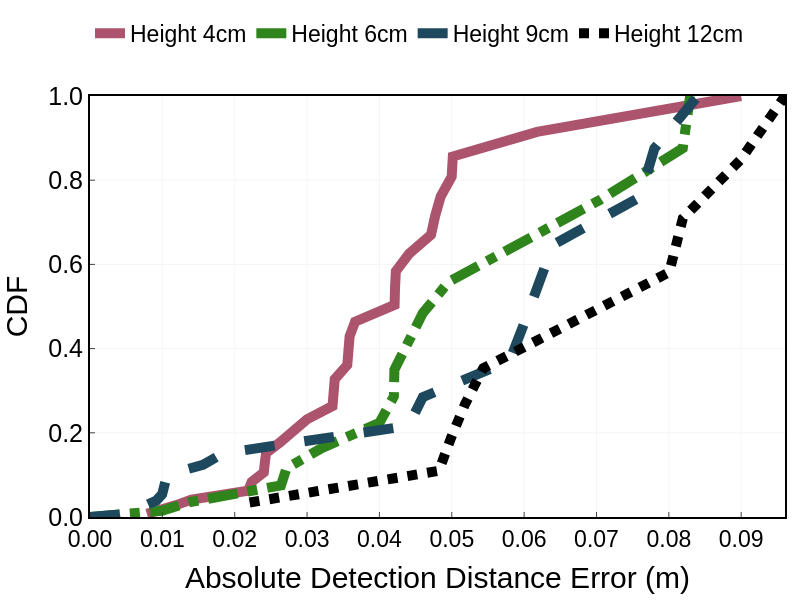}%
        \label{Performance:1d}
    }\hfill
    \subfloat[CDF of Different Angles]{%
        \includegraphics[width=0.325\linewidth]{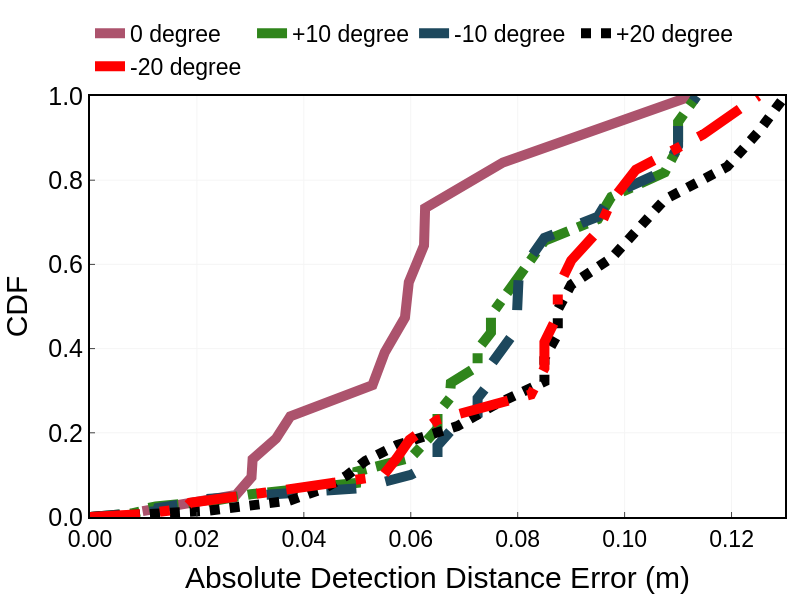}%
        \label{Performance:1e}
    }\hfill
    \subfloat[Errors of Different Sets of Materials]{%
        \includegraphics[width=0.325\linewidth]{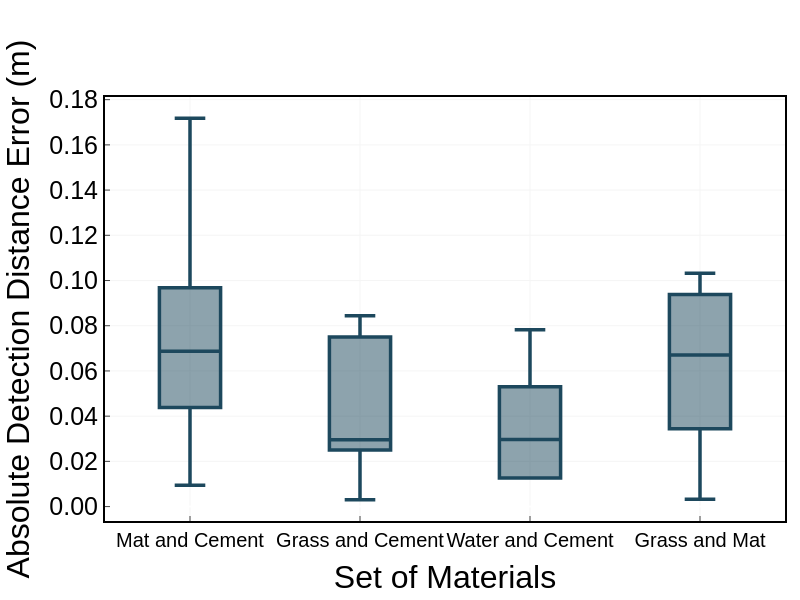}%
        \label{Performance:1f}
    }

    \vspace{-0.2cm}
    \subfloat[Efficiency of DF Loss]{%
        \includegraphics[width=0.325\linewidth]{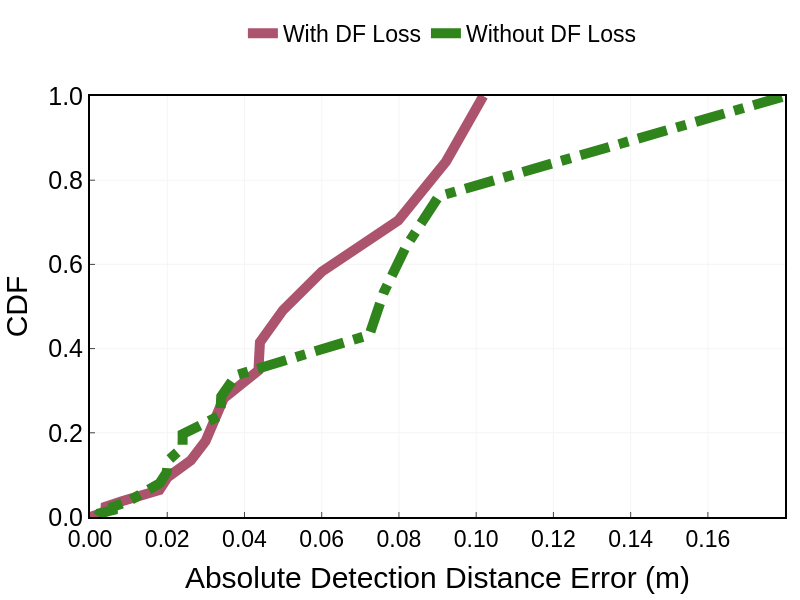}%
        \label{Performance:1g}
    }\hfill
    \subfloat[Impact of Time Precision]{%
        \includegraphics[width=0.325\linewidth]{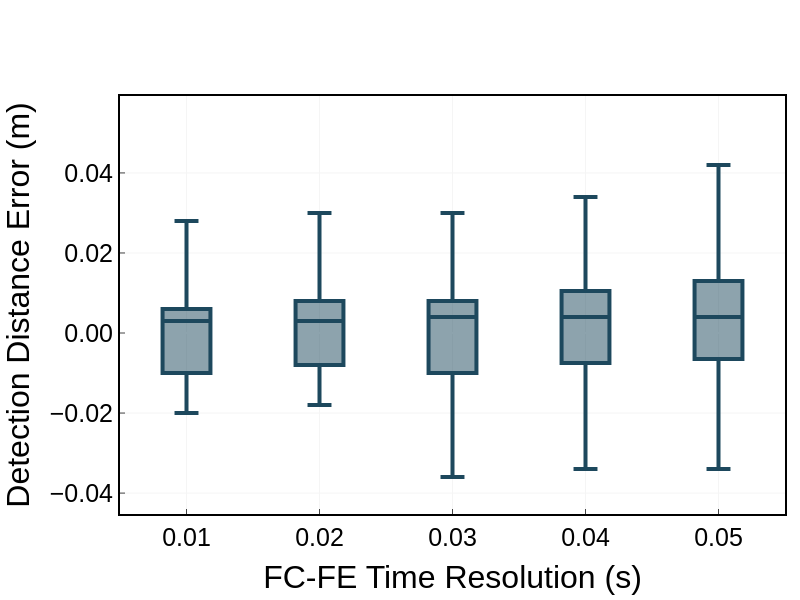}%
        \label{Performance:1h}
    }\hfill
    \subfloat[Impact of NN Input Size]{%
        \includegraphics[width=0.325\linewidth]{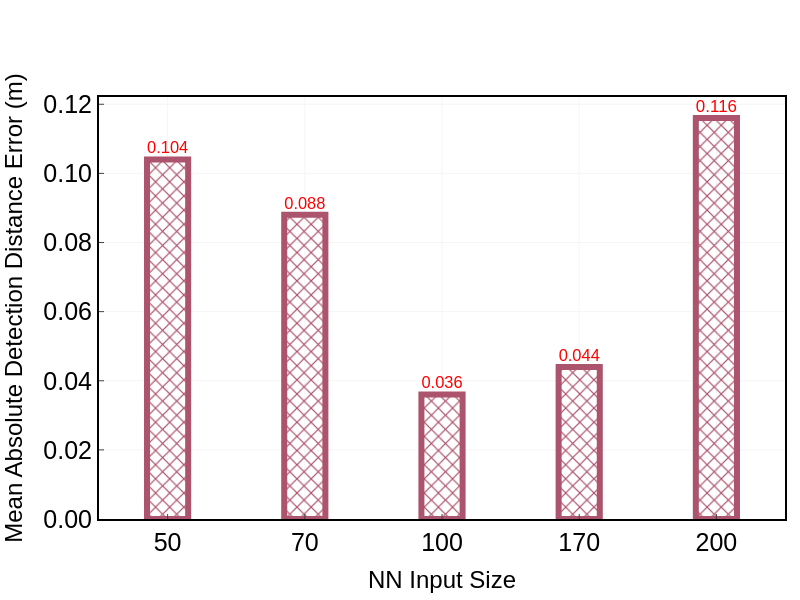}%
        \label{Performance:1i}
    }

    \caption{Performance. (a) The absolute detection errors in 30 standard flights for the height edge detection. (b) The absolute detection errors in 30 standard flights for the material edge detection. (c) Comparison with the baseline on different materials. (d) The detection errors with different heights to the surface. (e) The detection errors with different angles to the edge. (f) The comparison among different sets of materials. (g) The efficiency of DF Loss Function. (h) Influence of Time Resolution. (i) The performance influence of NN input layer size.}
    \label{Performance}
    \vspace{-0.2cm}
\end{figure*}

\subsection{Experiment Setup}

\textbf{Parameter Setup and Dataset: }
We measured Crazyflie's mass $m$, propeller radius $D$, the distance between rotor axes $d$, thrust constant $C_{T}$, and air pressure $\rho$. The drone is shown in Fig.~\ref{hardware}\subref{hardware:2a}. Also, we tested the thrust constant $k_{T}$ from the real world using the relationship $k_{T} = C_{T} \rho D^4$. The data sampling rate is 100Hz for acceleration rate, angular acceleration rate, and motor signals. For STFT, we set the window size and overlap size to 199. In FR-CFAR, we configure the following parameters: $P_{FA}=10^{-6}$, a leading window size of 50, and a guard cell size of 15. \waittoadd{For the dataset, we arranged the testbed of the height discontinuity platform, which is a drone landing platform, and several surfaces of different materials to collect the training dataset. We collect 200 flights for height and 300 flights for material edge detection. Each flight produces more than ten thousand training samples, so the total training samples are more than 5 million.}

 \textbf{Network Training:}
\ The proposed network is implemented in TensorFlow. We trained the models for 200 epochs with a fixed batch size of 32. The input window size is 100, corresponding to a total sequence length of 100 timestamps. The output represents the probability of the drone either reaching or not reaching an edge and is encoded using one-hot encoding, resulting in an output size of 2.

\textbf{Evaluation Metrics and Ground Truth:}
The performance of our system is evaluated using the \textbf{detection distance error}, which measures the discrepancy between the detected edge and the actual edge. The distance error is calculated by converting the time error of detection using the drone's velocity. \majorrevise{We align the ground truth of the precise edge location with the IMU and motor signals by calculating the distance between the drone and the landmark representing the edge. This landmark, fixed and carefully aligned by us, is recorded by the drone's multi-range deck (shown in Fig.~\ref{visionperformance}\subref{Performance:3a}), which measures the distance between the drone and the objects from one of four directions. Therefore, the sudden changes of the range sensor from one particular direction (e.g., right side) indicates the ground truth of edge occurs.}

\textbf{Baseline:}
\ We tested our system against a low computational cost frequency spectrum transform method. It first transforms the raw data using STFT and accumulates the power at every timestamp. Then, it calculates the sum of correlations using Pearson Correlation Coefficient~\cite{zhang2023rf}, and the target detected edge is at the point when the second-order derivative is equaled to zero.

\subsection{Overall Performance}
To evaluate the system's overall performance, we examined the accuracy of edge detection for abrupt height discontinuities and material interface transitions across 30 test groups. Fig.~\ref{Performance}\subref{Performance:1a} and  \ref{Performance}\subref{Performance:1b} depict the edge detection distance errors for the cases. 
The positive values stand for the early detection before the drone exactly passes above the edge, while negative values present the lagging detection.
Notably, abrupt height discontinuities can be regarded as one form of material interface transition, where one side of the surface is air. After analysis, the mean absolute detection distance error for abrupt height discontinuities was $0.0295$m, while it was $0.0362$m for material interface transitions.

We compare the edge detection errors of our method with the baseline. Our method outperforms the baseline in all sets of detection. The mean absolute error is 0.051 m, much lower than the baseline at 0.364m, with $86\%$ performance improvement. This is attributed to the characteristics of the ground effect refined by our system's design, so without profiling the proper characteristics, the baseline can only achieve coarse detection.

\subsection{Case Study: Abrupt Height Discontinuities}
\label{sec:7.3}
\majorrevise{When testing the edge detection performance for abrupt height discontinuities, we employed a platform constructed from an acrylic surface platform as illustrated in Fig.~\ref{hardware}\subref{hardware:2b}, over which the drone flew horizontally. Due to the platform's height elevation above the ground, the moment when the drone flies over the platform represents a scenario of abrupt height discontinuity at the edge. Such a platform can be considered a platform for drones to land or charge in application scenarios. 
As the drone flew above the platform, the ground effect on the aircraft intensified, leading to increased disturbance forces. 
Furthermore, we tested the system's edge detection robustness by varying the drone's height above the ground and the angle at which it approached the edge during horizontal flight.

The height is defined as the vertical distance above the drone to the platform surface when sweeping past it. As depicted in Fig.~\ref{Performance}\subref{Performance:1d}, detection accuracy decreases along with the height increases. At a height of 4cm, nearly 80\% of the errors fall within 0.05m, while with the heights of 9cm and 12cm, less than 30\% of the tests fall within this range. This trend is attributed to the increased altitude, which weakens the ground effect and the disturbance. The weaker the effect brings to the drone, the more challenging it is for detection.

We also explored the influence of the transition angle to the edge. The horizontal angle is defined as the angle between the drone’s horizontal flight path and the horizontal normal to the edge of the platform. Fig.~\ref{Performance}\subref{Performance:1e} illustrates that detection accuracy decreases when the angle increases. This may be because the asymmetry of the angle enhances the dynamics of drone's attitude, with obscure features. The overlap between the curves for +10 and -10 degrees, as well as +20 and -20 degrees, suggests that the magnitude of the horizontal angle significantly impacts the performance of detection accuracy, not the horizontal direction, due to the symmetric design of the drone. }

\subsection{Case Study: Material Interface Transitions}
\begin{figure}[t]
\vspace{-1.1em}
 \centering
    \subfloat[Lightweight Drone Hardware]{%
        \includegraphics[height=0.47\columnwidth, angle=-90]{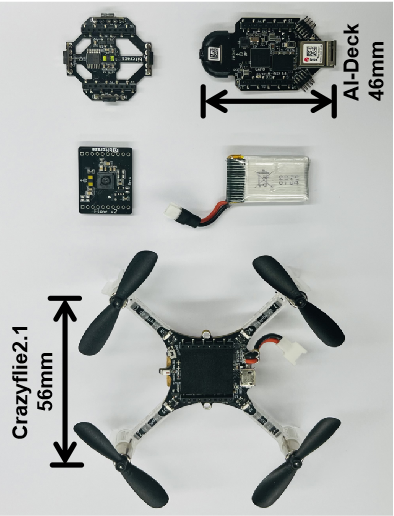}%
        \label{hardware:2a}
    }\hfill
    \subfloat[Height Detection Platform]{%
        \includegraphics[height=0.47\columnwidth, angle=-90]{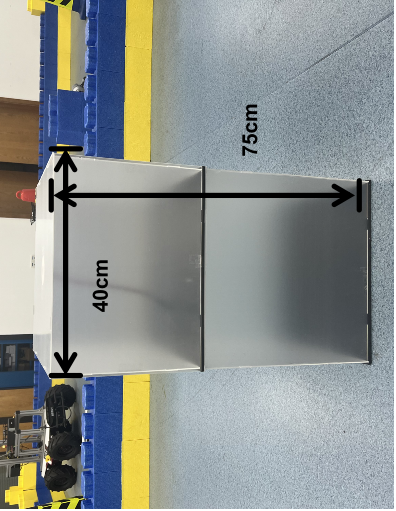}%
        \label{hardware:2b}
    }
\vspace{-0.5em}
    \subfloat[Mat and Cement]{%
        \includegraphics[height=0.47\columnwidth, angle=-90]{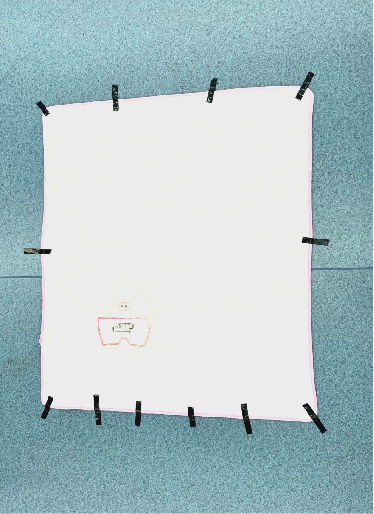}%
        \label{hardware:2c}
    }\hfill
    \subfloat[Artificial Grass and Cement]{%
        \includegraphics[height=0.47\columnwidth, angle=-90]{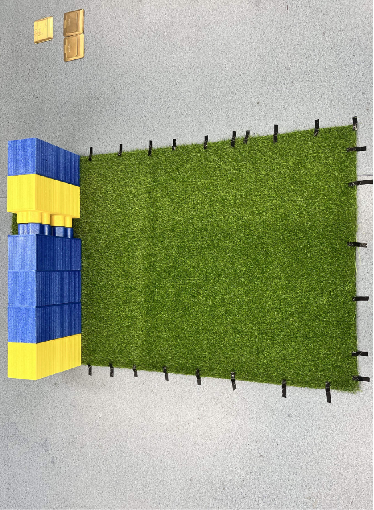}%
        \label{hardware:2d}
    }
\vspace{-0.5em}
    \subfloat[Water and Cement]{%
        \includegraphics[height=0.47\columnwidth, angle=-90]{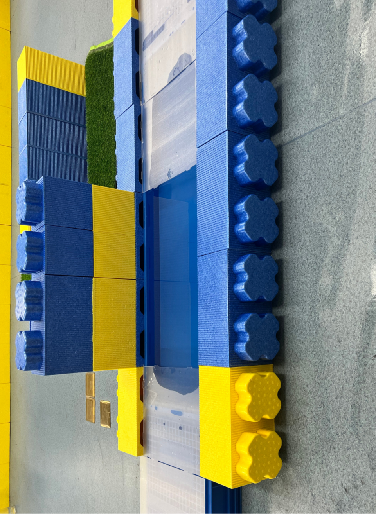}%
        \label{hardware:2e}
    }\hfill
    \subfloat[Artificial Grass and Mat]{%
        \includegraphics[height=0.47\columnwidth, angle=-90]{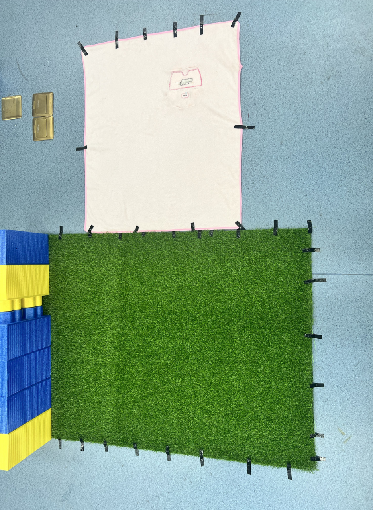}%
        \label{hardware:2f}
    }

    \caption{Hardware and testbeds of experiments for dataset collection and infield experiments.}
    \label{hardware}
\end{figure}

We conducted four sets of experiments to test the edge detection performance for Material Interface Transitions. These included edges between "mat" and "cement", "artificial grass" and "cement", "water" and "cement", and "artificial grass" and "mat" (Fig.~\ref{hardware}\subref{hardware:2c}-\subref{hardware:2f}).
According to Fig.~\ref{Performance}\subref{Performance:1f}, the edge detection error is the largest and most widely ranged for the "mat" and "cement" edges, with a median error of 0.070 m and a maximum error of 0.170 m. It is slightly lower but relatively stable for the "artificial grass" and "cement" edges. The "water" and "cement" edges exhibit the smallest and most concentrated detection errors. Approximately $75\%$ of the data falls within the range of 0.01 m to 0.05 m. 
The edge detection error between "artificial grass" and "mat" falls between that of "mat" and "cement" and that of "artificial grass" and "cement".

The surface natures of different materials
generate different ground effects in terms of intensity, frequency, and directional components, leading to different drone behaviors. If the discrepancy is more obvious, the detection will be relatively more accurate because it is easier to capture the edge-induced change (with higher SNR) and vice versa. For example, the artificial grass surface's pronounced texture and varying height compared to the smooth cement surface contribute to this distinction in ground effect, making the edge more discernible for the artificial grass.

\subsection{Impact of System Components}

\majorrevise{\textbf{Effectiveness of DF Loss Function: }To evaluate the effectiveness of the Disturbance Force-Informed (DF) Loss Function, we conducted a comparative analysis between models trained with and without the DF loss. As depicted in Fig.~\ref{Performance}\subref{Performance:1g}, the introduction of the DF loss led to a significant reduction in distance detection errors, with the majority of errors decreasing from 0.18 m to about 0.10 m. This improvement demonstrates a substantial accuracy improvement, reaching 44\% compared to the model trained without the DF loss. When the CDF curve of errors with DF loss reaches 100\% at around 0.10 m, the curve without DF loss just reaches under 80\%. The difference in the performance comes from the physical knowledge and fine-grained bias from DF loss, particularly when the model is trained using a small amount of flight data. Additionally, the network convergence speed has been expedited. The training process of the model with DF loss converged after around 190 epochs, while without it the model needs more than 260 epochs, improved by nearly 27\%.}

\textbf{FC-FE Time Resolution: }We also investigated the impact of time resolution in the Fluctuation Components Feature Extraction (FC-FE) process. The resolution can be adjusted by modifying the window size and overlap size. As depicted in Fig.~\ref{Performance}\subref{Performance:1h}, higher resolutions enhanced the detection performance, resulting in lower distance errors.

\textbf{NN Input Window Size: }To assess the influence of different input window sizes, we conducted experiments on a standardized dataset. As illustrated in Fig.~\ref{Performance}\subref{Performance:1i}, the input size ranging from 100 to 170 exhibited the best results. Considering the trade-off between model performance and computational complexity, we chose a window size of 100 as it strikes a balance, maintaining satisfactory performance levels.

\subsection{Model Optimization and Efficiency}

\begin{figure}[t]
\centering
\includegraphics[width=0.85\linewidth]{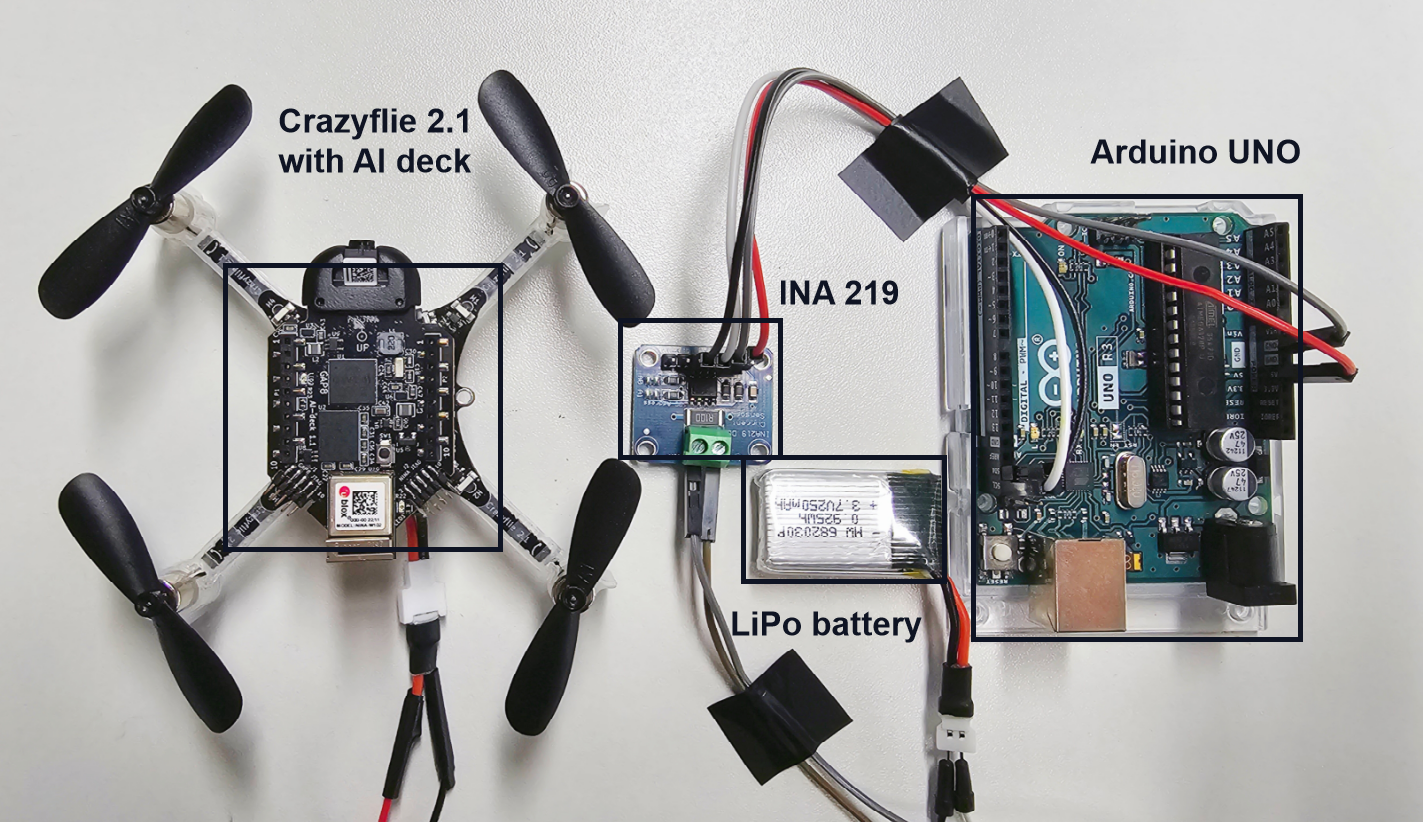}
\caption{Power consumption measurement Setup using INA 219 and Arduino.}
\label{power}
\end{figure}


\waittoadd{
After pruning and quantizing our NN model, we observed a reduction in model size from 141,837 Bytes to 14,375 Bytes. Additionally, the inference time was reduced from 0.00271s to 0.00110s. However, the model's accuracy on the training and test sets decreased from $97\%$ to $86\%$.  
Notably, the accuracy reduction is deemed acceptable the model size has been reduced by nearly 10-fold, while the inference speed has increased by approximately 2.5 times. Although this results in an $11\%$ accuracy drop to $86\%$, the trade-off is acceptable given the system's focus on low power consumption and real-time performance. Additionally, we endeavored to minimize the pruning of the neural network in an effort to enhance detection accuracy. However, when the detection accuracy reaches approximately $92\%$, the model's weight size significantly increases, occupying nearly 123,921 Bytes of storage space while maintaining an inference speed comparable to that of the original model. This phenomenon arises due to the inherently non-linear relationship between detection performance and model weight size. Given this trade-off, we concluded that it is advantageous to accept a $6\%$ reduction in detection accuracy (from $92\%$ to $86\%$) in exchange for a substantial improvement in efficiency. Specifically, the model weight size is reduced from 123,921 Bytes to 14,375 Bytes, representing a significant gain in both storage efficiency and potential computational performance. Overall, the neural network pruning reduces the model size and computational complexity significantly, enabling real-time deployment on resource-constrained drones, and the pruned model still outperforms the baseline method.

}

To demonstrate the performance and energy efficiency of AirTouch, we conducted power consumption experiments by deploying our modules on Crazyflie's STM32 and the AI deck's GAP8 processor. As shown in Fig.~\ref{power}, we utilized the INA 219 and Arduino to measure the power consumption of the nano-quadrotor with AI deck during the deployment of our system on board. The INA 219 is a current sensor that provides accurate current and voltage measurements.

\renewcommand{\arraystretch}{1}
\begin{table}[t]
\vspace{-1em}
\caption{System power consumption}
\centering
\begin{tabular}{lc}
\toprule
Module & Power (mW)\\
\midrule
Basic drone module power & 78 \\
Fluctuation Components & 6 \\
Feature Extraction & \\
Cascaded Cross-Spectrum & 2\\
Feature Fusion &  \\
Aerodynamics-Informed Double  & 2\\
Phase Physical Filter & \\
Physical Knowledge-aided Network & 33 \\
Total Power & 121 \\
\bottomrule
\end{tabular}
\label{t1}
\end{table}

\waittoadd{It is a direct measurement of power consumption for each component in Table.\ref{t1}, which serves as a proxy for understanding the system's energy efficiency. This data is used to estimate the potential impact on flight time based on their specific drone configurations and mission requirements. For instance, with a drone's total power budget and battery capacity known, one can calculate the expected reduction in flight time using the power consumption figures we have reported. It allows for a standardized comparison of energy efficiency across different systems and provides a foundation for further optimizations to minimize the impact on flight time. Table.\ref{t1} shows that the network module consumes the most power, 33 mW, contributing significantly to the total system power consumption of 121 mW. However, the power consumption for driving the drone's propellers during flight is approximately 3000 mW. Therefore, the energy consumption of our proposed system is negligible.}

\waittoadd{
\subsection{Environmental Wind Influences}

In our experimental setup, we validated the performance of AirTouch in both indoor and outdoor environments. For the indoor experiments, we simulated real-world wind conditions by utilizing air conditioners and opening/closing doors and windows, thereby introducing varying wind conditions into the test area. These experiments allowed us to evaluate the robustness of our system in environments with wind disturbances, which are commonly encountered in indoor settings. The results demonstrated that AirTouch can effectively detect edges even in the presence of these wind factors, as the system focuses on detecting sudden changes in ground effect caused by edges rather than the absolute strength of the airflow. This is supported by our analysis of the drone's aerodynamics and flight control mechanisms. To validate this concept, we implement a constant wind generated by a fan on the left and front sides, respectively. The result demonstrates that the edge detection performance of AirTouch remains similar with wind speeds from 0 m/s, 0.1 m/s, 0.2 m/s, and 0.3 m/s. The results demonstrate that the proposed system exhibits robustness and stability when subjected to gentle winds characterized by relatively constant direction and strength.

From a theoretical perspective, the AirTouch system relies on detecting abrupt changes in the ground effect, which are indicative of edges. Relatively steady and constant wind conditions do not significantly impact the system's performance because the detection mechanism is based on identifying sudden changes in the drone's flight dynamics. Our analysis of the drone's aerodynamics and flight control mechanisms supports this, showing that the system can maintain reliable edge detection performance under such conditions.}

\majorrevise{
\subsection{Comparison With Vision}
AirTouch is a compact and complementary solution for edge detection and it shows various apparent advantages in several aspects over vision-based methods.

\renewcommand{\arraystretch}{1.2}
\begin{table}[t]
\caption{\majorrevise{Resource Efficiency}}
\centering
\begin{tabular}{c c c c}
\toprule
Items & Vision-based & AirTouch & Cost ratio\\
\midrule
FLOPs & 3900 million & 0.28 million & 0.07$\%$ \\
NN parameter size & 58KBytes	& 13KBytes & $ 22\%$ \\
Training time & 30min	& 13min & $ 43\%$ \\
Memory per frame & Millions of bits & 32Kbits & $< 1\%$ \\
\bottomrule
\end{tabular}
\label{t2}
\vspace{-1.5em}
\end{table}

\textbf{Resource efficiency: }AirTouch demonstrates a significant reduction in computational and memory requirements, as is shown in Table. \ref{t2}. According to our tests, AirTouch requires only 0.28 million FLOPs and 13KB of neural network parameters, amounting to 0.007\% of the FLOPs and 22\% of the parameter size found in the state-of-the-art lightweight vision neural network (i.e., FLOPs: 3900 million, parameter size: 58 KB)~\cite{soria2023tiny}. Although a low-resolution camera (320x320 grayscale) processes millions of bits per frame, AirTouch uses only 32kbits—substantially less. Additionally, our use of advanced techniques like factorized optimization and group convolution holds promise for further reducing its resource footprint. 

\textbf{Accuracy performance: }We compare the edge detection accuracy of AirTouch with the vision-based edge detection method. We choose TEED~\cite{soria2023tiny} as the compared vision method, which is the state-of-the-art lightweight CNN to tackle edge detection tasks, specifically designed for simplification, efficiency, and generalization. The camera used for data collection and detection test is the Himax HM01B0, an ultra-low power 320×320 monochrome camera, equipped on the Crazyflie's AI deck. 

\begin{figure*}[htbp]
    \centering
    \vspace{-0.5cm}
    \subfloat[\majorrevise{Range sensor for ground truth and camera for comparison with vision}]{%
        
        \includegraphics[width=0.325\linewidth]{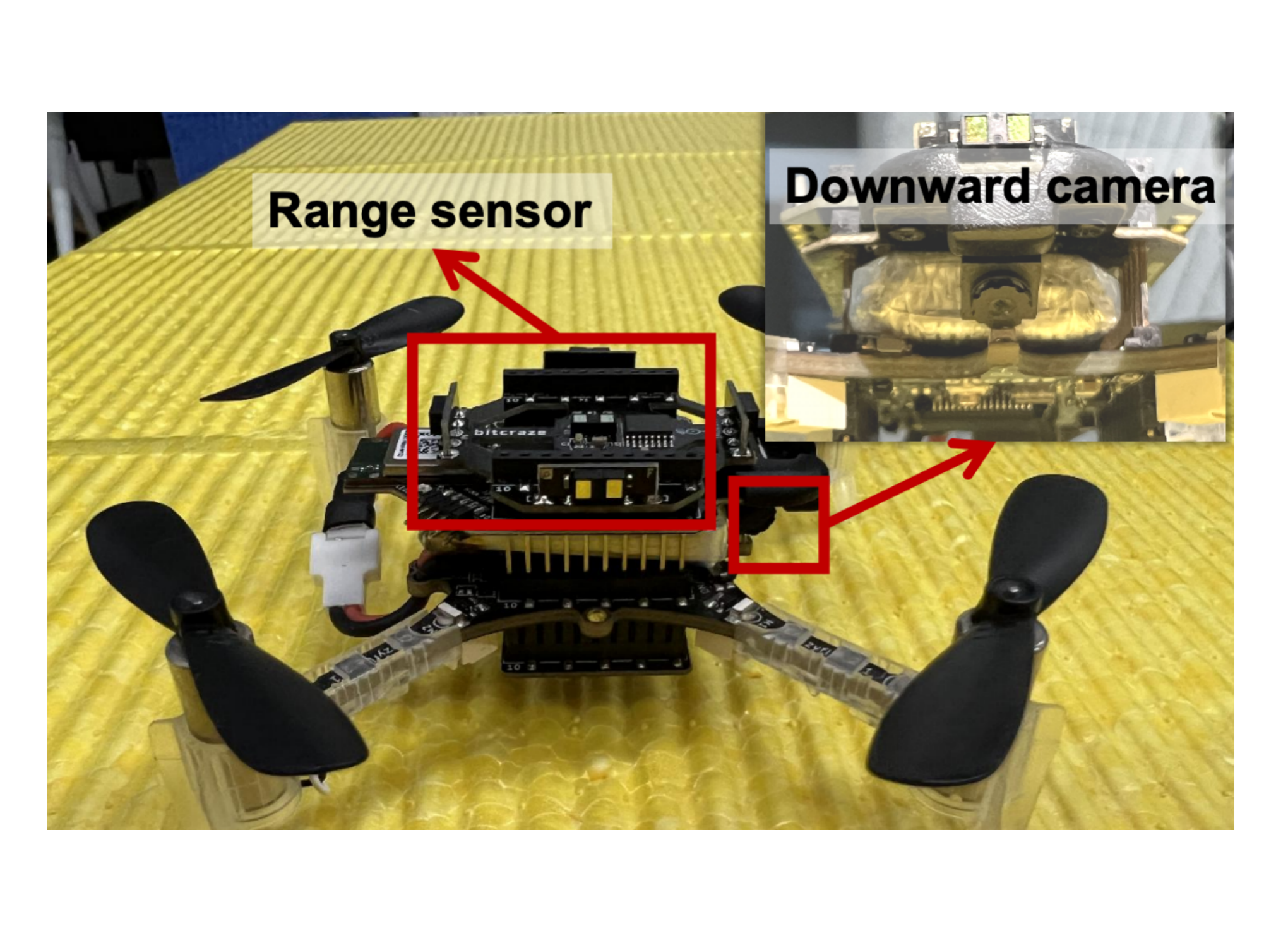}%
        \label{Performance:3a}
    }\hfill
    \subfloat[\majorrevise{Errors in different luminance}]{%
        \includegraphics[width=0.325\linewidth]{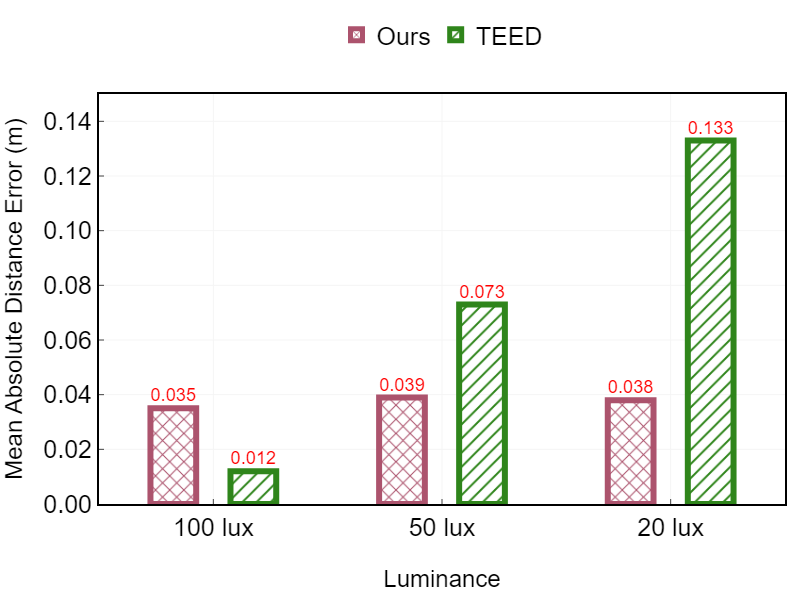}%
        \label{Performance:3b}
    }\hfill
    \subfloat[\majorrevise{Errors with different NN parameter size}]{%
        \includegraphics[width=0.325\linewidth]{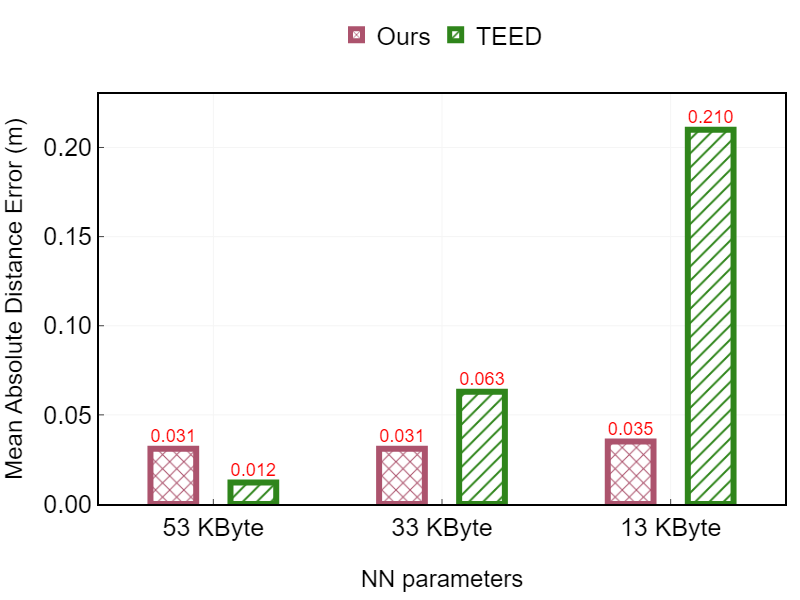}%
        \label{Performance:3c}
    }

    \caption{\majorrevise{Comparison with vision. (a) The multi-range sensor is to detect milestones of distance change to record the ground truth and the downward camera is for visual edge detection. (b) The absolute detection errors in 20 standard flights for the height edge detection with vision-based method under different light conditions. (c) The absolute detection errors in 20 standard flights for the height edge detection compared with vision-based method under different neural network parameters.}}
    \label{visionperformance}
    \vspace{-0.2cm}
\end{figure*}

We test the height change edge detection performance of both AirTouch and TEED in scenarios of different light conditions. We use the default configuration for the luminance test, with TEED neural network parameter size of 53 KB and AirTouch parameter size of 13KB, distance to surface of 4cm, and angle to the edge of 0 degrees. With the degradation of luminance, as shown in Fig.~\ref{visionperformance}\subref{Performance:3b}, our system maintains the same detection accuracy with mean absolute distance errors around 0.035m. However, mean detection errors of TEED increase while the light is getting poorer. At the luminance of 100 lux, vision-based TEED outperforms AirTouch with a mean error of around 0.012m. However, at 50 lux and 20 lux luminance, the accuracy decreases to levels worse than AirTouch's in low-light conditions. This result mainly comes from the fact that the performance of edge detection diminishes due to a reduced signal-to-noise ratio, making it harder for vision algorithms to accurately identify and extract clear edge contours or distinguish between true edges and noise. The subtle gradient information in the image becomes less pronounced, complicating the detection of edge positions in the image and then affecting the edge detection accuracy in the physical world when the drone flies past the boundaries. Also, due to the hardware limitations, the camera can only provide around 6 frames in one second so the detection resolution may deter the accuracy.

In addition, we set different layers and units of two neural networks to compare their performance under different sizes of neural network parameters. From Fig.~\ref{visionperformance}\subref{Performance:3c}, we can find that when the size of NN parameters of TEED decreases from its default 53KB, the mean absolute detection error increases, which means that the dropped layers and units contain a considerable amount of necessary representational capability for edge detection. In contrast, the accuracy of AirTouch does not increase a lot when more parameters are involved. This may be because feature extraction, feature fusion, and the physical filter have already manifested the edge information and simplified the detection process. Therefore, a more powerful characterizing capability with a large parameter size may not help a lot with detection performance.


\textbf{Complement as a backup:} While AirTouch is explicitly crafted to be a complement, not a substitute, for current methodologies, its significance lies in its ability to address the shortcomings of visual modalities under challenging circumstances such as low-light environments or scenarios with similar color and texture patterns where traditional methods falter. By leveraging its capability to discern edges through unique physical attributes, AirTouch emerges as a valuable augmentation to existing systems, serving not only as an additional resource but also as a crucial contingency plan during instances of critical system malfunctions.

}
\section{Discussion}
\label{sec:discussion}
In this section, we provide explanations of some concerns.

\waittoadd{
\subsection{Sensing Range}
In \S \ref{sec:7.3}, the results show the sensing effectiveness degraded significantly above 12 cm. In fact, the sensing range is extendable with larger, heavier drones. The sensing range is closely tied to the drone's physical characteristics and operating conditions. From the theoretical aspect, according to recent studies of aerodynamics and ground effect~\cite{conyers2018empirical, matus2021ground}, larger drones, longer propellers, and heavier weights generate more intense airflows and GE, even when flying at the same altitude. As a result, the GE will dissipate over a greater distance for detection, thereby extending the sensing range. Our prior experiments showed that a drone weighing 2.5 kg with a diagonal span of 450 mm and 100 mm propellers could detect ground effect from approximately 3 meters above the ground, indicating the feasibility of range extension through adjustments to these parameters. Although the current experiment results show the detection range is limited, they have validated the functionality of the proposed new sensing modality for material and height edge detection, confirming the core intuition and theoretical framework of this work.

In the future, it is a promising research direction to systematically investigate the impact of propeller size and drone weight on the sensing range. By varying these parameters while keeping other factors constant, we aim to measure the corresponding changes in detection range. This research will deepen the understanding of how drone design influences sensing capabilities and will guide the development of a more versatile system adaptable to diverse drones and operational requirements. The objective is to enhance the practical applicability of the proposed approach, making it a more robust solution for real-world edge detection tasks.
}

\waittoadd{
\subsection{More Challenging Winds}
Our testing has primarily been conducted in controlled indoor environments, and actual outdoor wind conditions can be more complex and variable. Conducting comprehensive outdoor experiments with an anemometer to measure wind speed and direction, alongside detection accuracy, is a valuable next step to further validate the system's robustness and applicability in diverse environments. In future work, we plan to expand our experiments to include more extensive outdoor testing. By using an anemometer to measure wind conditions and collecting data on how different wind speeds and directions affect the detection accuracy of our system, we aim to better understand the system's limitations and identify potential improvements needed for real-world outdoor applications. Additionally, we will explore enhancements to our algorithms and control strategies to better handle complex wind fields, ensuring that AirTouch can maintain reliable performance across a wider range of conditions.}

\waittoadd{
\subsection{Complex Edge Scenarios}
Currently, AirTouch is proficient in detecting quadrilateral edges but may face challenges when dealing with more complex shapes or edges between surfaces of similar materials. To overcome these limitations, future research should prioritize several key areas. For instance, developing advanced flight path planning techniques could optimize sensing trajectories, thereby improving the detection of a wider variety of target geometries. Addressing the challenges posed by intricate sensing and control mechanisms will be crucial in enhancing the system’s adaptability. By optimizing flight paths, quadrotors can be guided along trajectories that maximize the capture of relevant airflow changes, even in complex environments. This not only improves detection accuracy but also reduces the computational load by focusing on the most informative regions.

Additionally, upgrading edge detection algorithms by training machine learning models on diverse datasets that include intricate shapes can further enhance detection compatibility. For scenarios that involve edges of similar materials, integrating supplementary sensing modalities, such as depth sensors or cameras, can provide additional contextual data, thereby increasing the overall success rate of detection. These advancements will not only refine the system's capabilities but also expand its applicability across various domains.

Moreover, future work can explore the fusion of multiple sensing modalities to create a more comprehensive perception system. Combining AirTouch with other sensor technologies could provide redundant information channels, making the system more robust to environmental variations and improving its ability to discern subtle differences between similar materials. Furthermore, the development of more sophisticated machine learning models, such as ensemble methods or multi-task learning frameworks, could help in capturing a broader range of features and improving the generalization capabilities of the edge detection system. By pursuing these research directions, AirTouch can be further advanced to handle complex real-world scenarios with greater reliability and precision.}





\waittoadd{
\subsection{Potential Applications}
The AirTouch system offers significant practical benefits for lightweight drones in various real-world applications. In disaster relief, AirTouch-equipped drones can quickly detect edges in debris fields, identify safe landing zones, and locate survivors trapped in complex terrains, thereby enhancing the efficiency and safety of rescue missions. For autonomous navigation, AirTouch enables drones to operate effectively in both indoor and outdoor environments by detecting terrain and obstacle edges, allowing for safe, collision-free flight even in GPS-denied areas. Additionally, AirTouch can be applied in environmental monitoring, agricultural production, industrial inspection, and logistics, where its ability to detect edges with minimal power consumption and computational resources makes it an ideal solution for enhancing situational awareness and operational capabilities. The system's reliability in low-light conditions and its insensitivity to surface properties further expand its applicability across diverse scenarios, making it a versatile tool for improving the performance and reliability of lightweight drones in precise and efficient sensing tasks.}

\vspace{-0.3cm}
\section{Conclusion}
\label{sec:10}
In conclusion, this paper introduces AirTouch, a proprioceptive sensing system that transforms the traditionally negative ground effect into a positive sensing modality for environmental edge detection. \majorrevise{By combining IMU and motor signals, AirTouch successfully captures and analyzes the ground effect in flight dynamics, providing a novel and effective approach for precise and efficient sensing tasks. The presented system exhibits notable performance in edge detection accuracy on lightweight drones while considering energy consumption and limited computational resources, and represents particular superiority over vision-based methods in some aspects. These achievements highlight the system's potential to enhance the capabilities of drones across diverse environments. Future research will concentrate on extending edge detection capabilities to mobile platforms, facilitating air-ground coordination and collaboration, which requires deeper collaboration of sensing and control.} Also, we will extend the AirTouch with multimodal fusion techniques to fully utilize its potential. As an initial work, this new sensing modality could benefit the community and inspire further research into its varied applications.

\section{Acknowledgement}
This paper was supported by the National Key R\&D program of China (2022YFC3300703), the Natural Science Foundation of China under Grant 62371269, Shenzhen 2022 Stabilization Support Program (WDZC20220811103500001) and Meituan.

\bibliographystyle{IEEEtran}
\bibliography{References}

\vspace{-1cm}
\begin{IEEEbiography}
[{\includegraphics[width=1.3in,height=1.30in, clip,keepaspectratio]{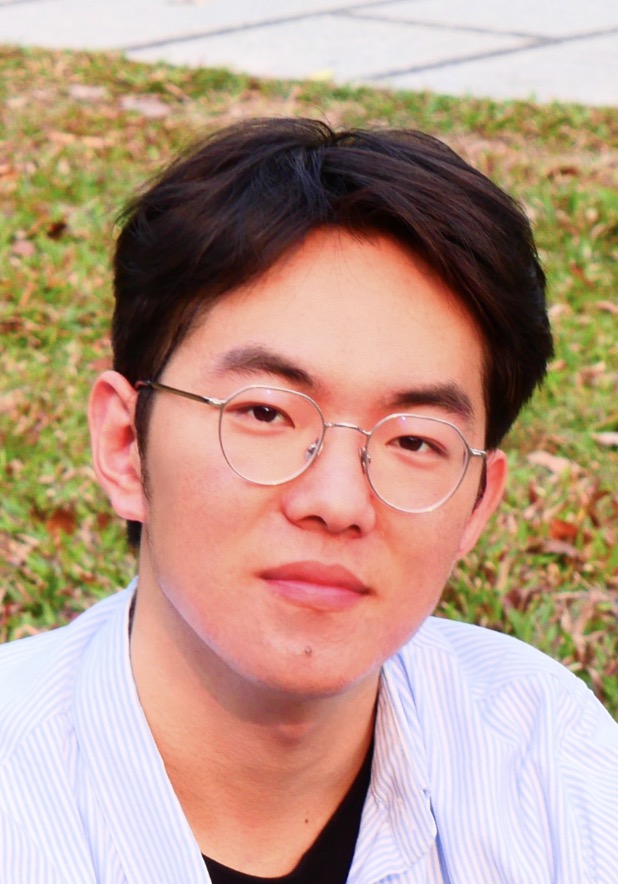}}]{Chenyu Zhao} is currently pursuing his M.Sc. degree at the Shenzhen International Graduate School, Tsinghua University, China. He received his B.E. degree of Electronic Information Science and Technology from Northwest University in China, and the B.S. degree in Electronic System Engineering from the University of Essex in UK, in 2022. His research interests include Mobile Computing, Cyber-Physical Systems, Embedded AI, Robotics, and Quadrotors.
\end{IEEEbiography}
\vspace{-1.2cm}

\begin{IEEEbiography}
[{\includegraphics[width=1.3in,height=1.25in, clip,keepaspectratio]{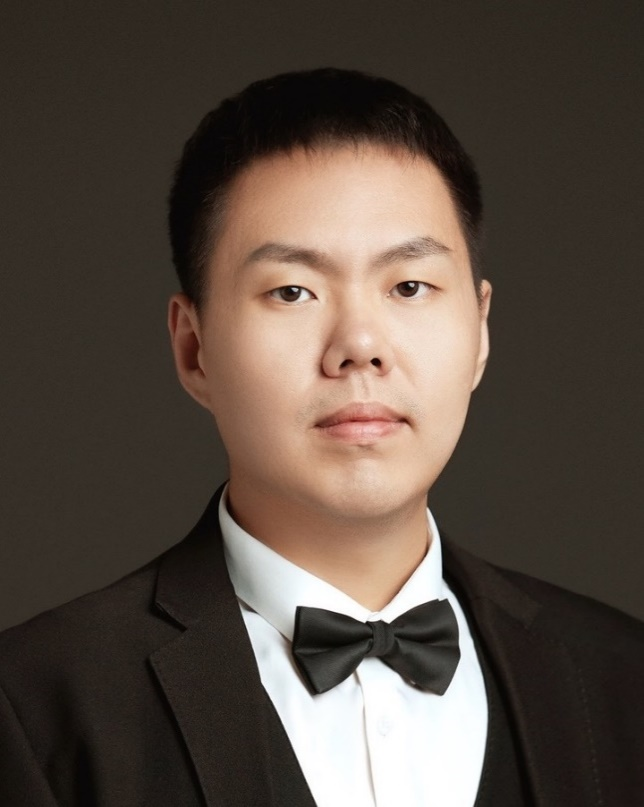}}]{Jingao Xu} is now a postdoctoral researcher in the Computer Science Department, Carnegie Mellon University. He received both his Ph.D. and B.E. degrees from Tsinghua University in 2022 and 2017, respectively. His research interests include Internet of Things, mobile computing, and edge computing.
\end{IEEEbiography}
\vspace{-1cm}

\begin{IEEEbiography}
[{\includegraphics[width=1.3in,height=1.25in,clip,keepaspectratio]{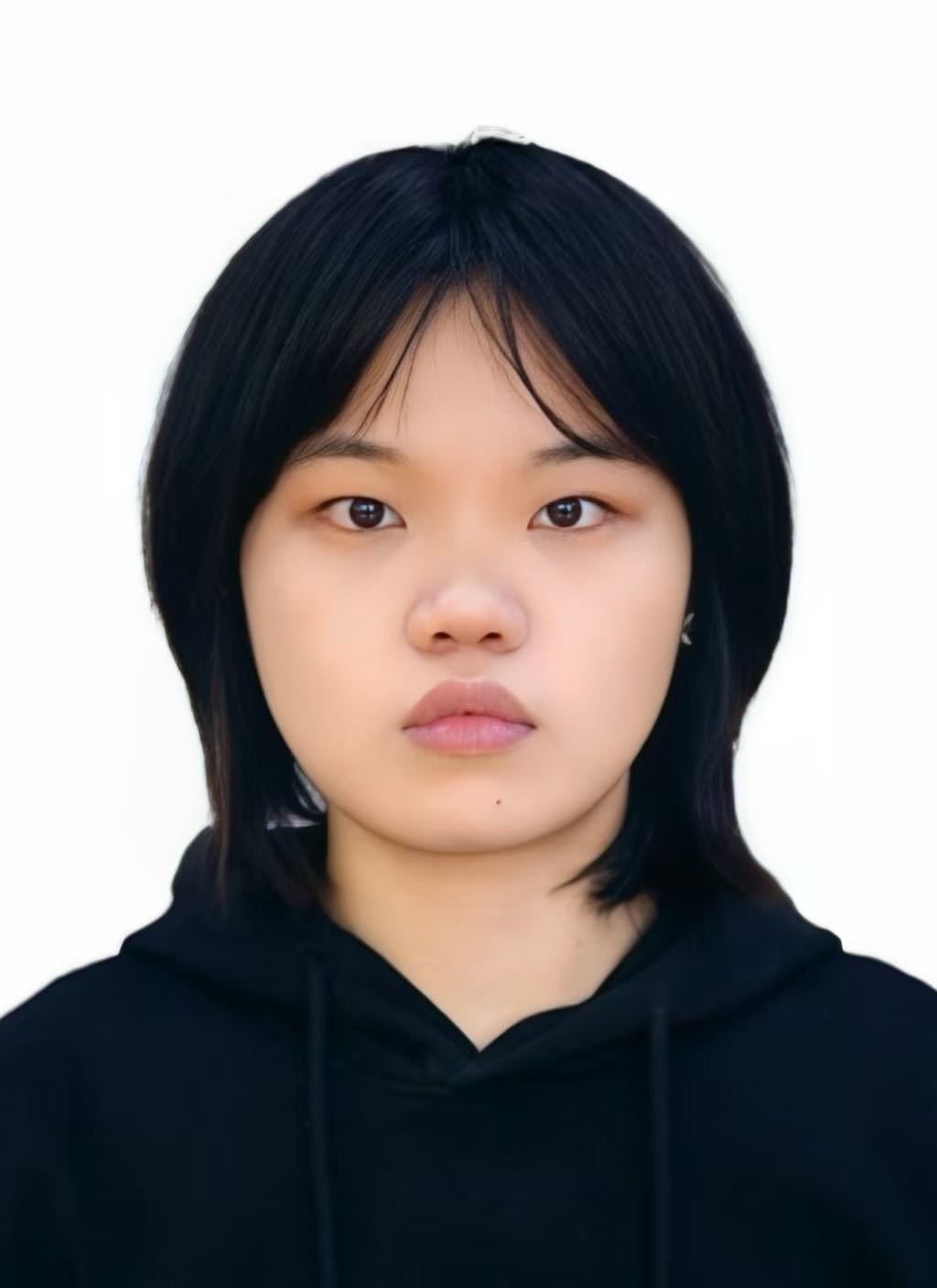}}]{Ciyu Ruan} received the B.E. degree from College of Intelligent Science, National University of Defense Technology, China, in 2023. She is currently pursuing the Master degree at the Shenzhen International Graduate School, Tsinghua University, China. Her research interests include mobile sensing, AIoT and Robotics.
\end{IEEEbiography}
\vspace{-1cm}

\begin{IEEEbiography}
[{\includegraphics[width=1.3in,height=1.25in,clip,keepaspectratio]{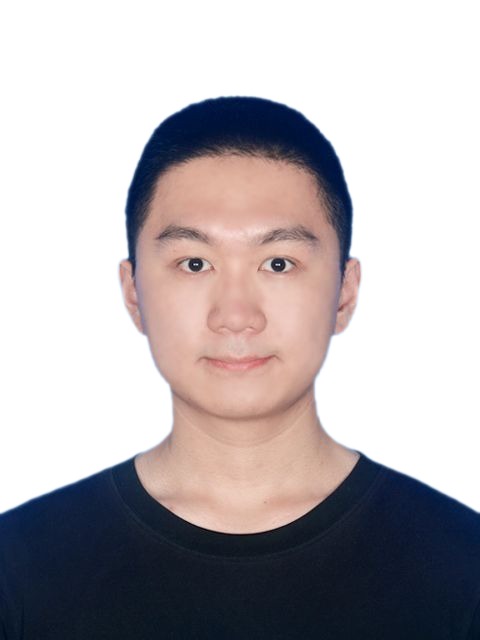}}]{Haoyang Wang} received the B.E. degree from the School of Computer Science and Engineering, Central South University, China, in 2022. He is currently pursuing the Ph.D. degree at the Shenzhen International Graduate School, Tsinghua University, China. His research interests include mobile computing, AIoT and distributed \& embedded AI.
\end{IEEEbiography}
\vspace{-1cm}

\begin{IEEEbiography}
[{\includegraphics[width=1.3in,height=1.25in,clip,keepaspectratio]{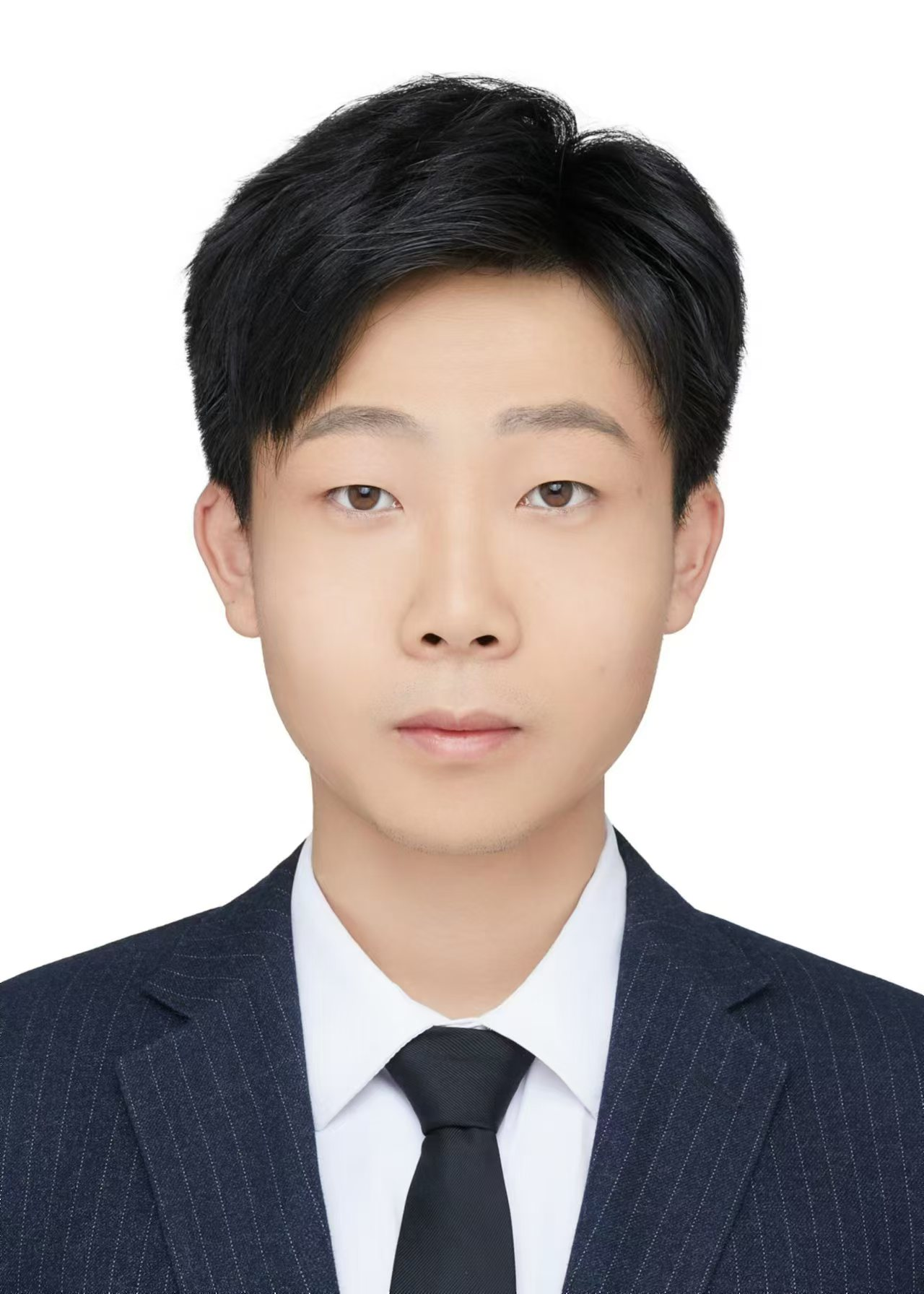}}]{Shengbo Wang} received a B.E. degree from the School of Instrumentation and Optoelectronic Engineering, Beihang University, China, in 2023. He is currently pursuing an M.Sc. degree at the Shenzhen International Graduate School, Tsinghua University, China. His research interests include mobile computing and edge computing.
\end{IEEEbiography}

\begin{IEEEbiography}
[{\includegraphics[width=1.3in,height=1.25in,clip,keepaspectratio]{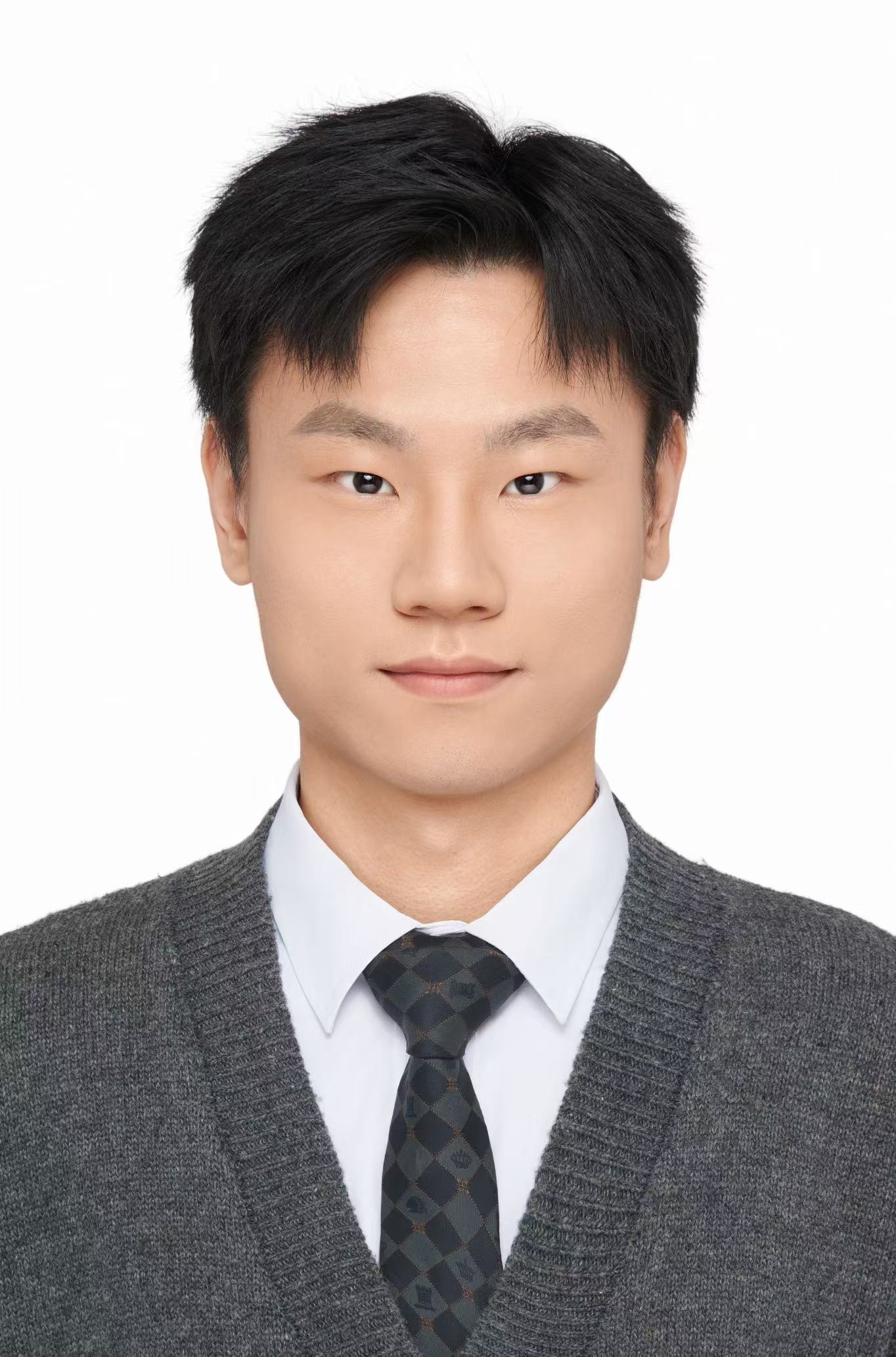}}]{Jiaqi Li} received the B.S. degree from the College of Artificial Intelligence and Big Data, University of Science and Technology of China. He is currently pursuing his Master's degree at the Shenzhen International Graduate School, Tsinghua University. His research interests include cyber-physical security and human-centered computing.
\end{IEEEbiography}

\begin{IEEEbiography}
[{\includegraphics[width=1.3in,height=1.25in,clip,keepaspectratio]{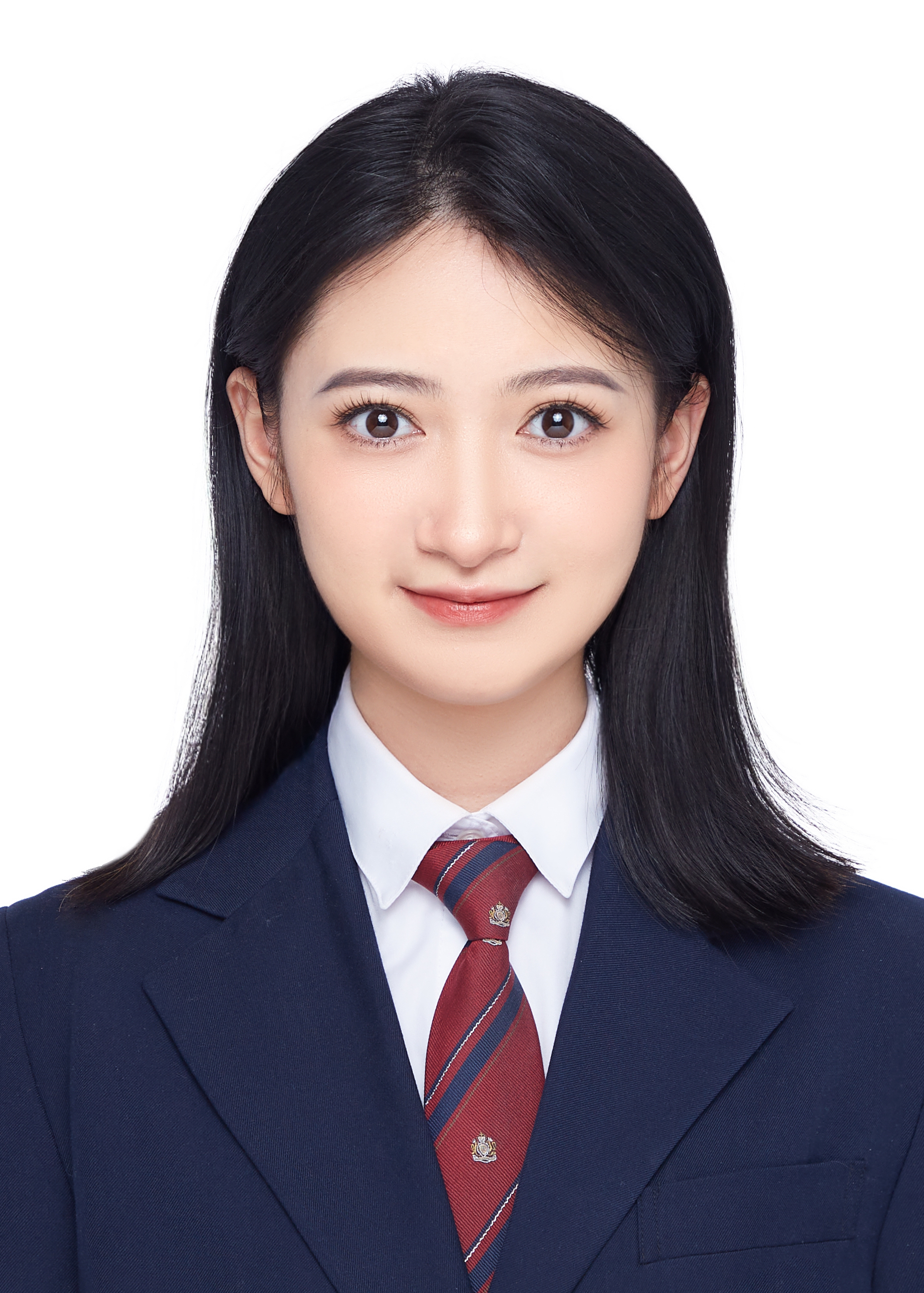}}]{Jirong Zha} received the B.S. and M.S. degrees from Beihang University, China, in 2020 and 2023, respectively. She is currently working toward the Ph.D. degree in data science and information technology from Tsinghua University, China. Her research interests include distributed state estimation, collaborative perception, machine learning, and multi-agent systems.
\end{IEEEbiography}

\begin{IEEEbiography}
[{\includegraphics[width=1.3in,height=1.25in,clip,keepaspectratio]{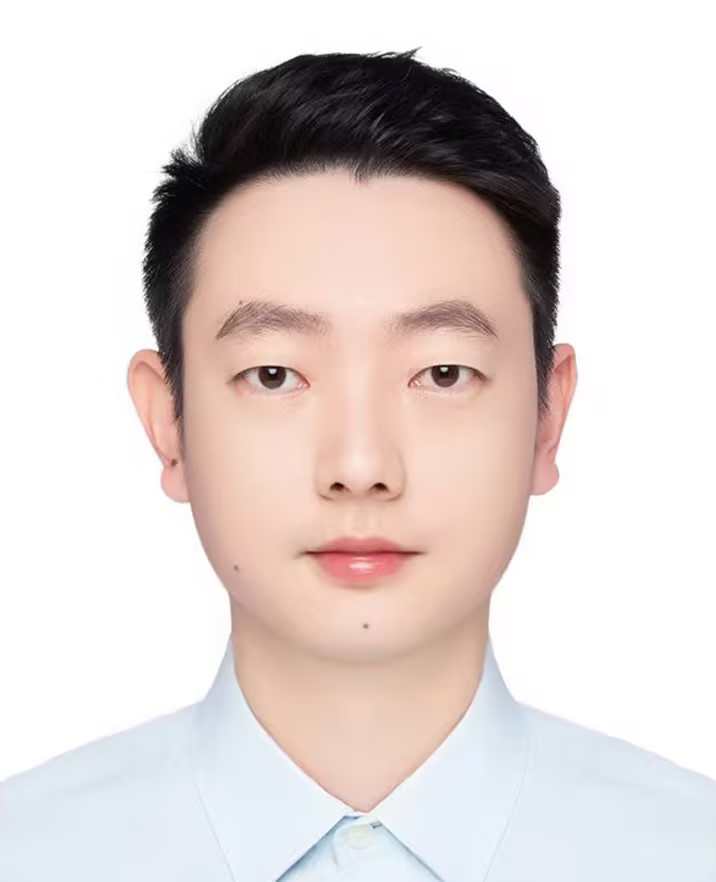}}]{Weijie Hong} obtained a bachelor's degree from South China University of Technology in 2009 and is currently pursuing a master's degree at Tsinghua University. He currently serves as Deputy Director of the Technology Department of Shenzhen Smart City Technology Development Group Co., Ltd., Technical Director of Shenzhen Smart City Communications Co., Ltd., and Director of Guangdong Engineering Technology Research Center for Multimodal Fusion Communication and IoT Sensing. 
\end{IEEEbiography}

\begin{IEEEbiography}
[{\includegraphics[width=1.3in,height=1.25in,clip,keepaspectratio]{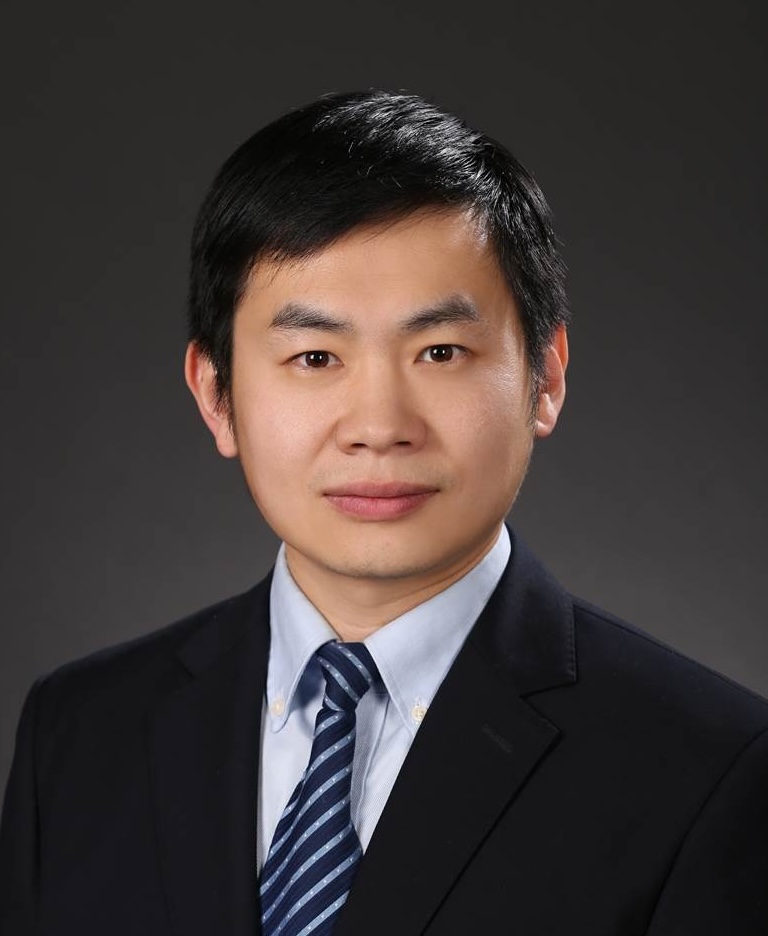}}]{Zheng Yang} is currently an assistant professor in School of Software and TNList, Tsinghua University, Beijing, China. He received his B.E. degree in the Department of Computer Science from Tsinghua University in 2006, and his Ph.D. degree in the Department of Computer Science and Engineering of Hong Kong University of Science and Technology in 2010. His research interests include Internet of Things, Industrial Internet, sensing and positioning, edge computing, etc. 
\end{IEEEbiography}

\begin{IEEEbiography}
[{\includegraphics[width=1.3in,height=1.25in,clip,keepaspectratio]{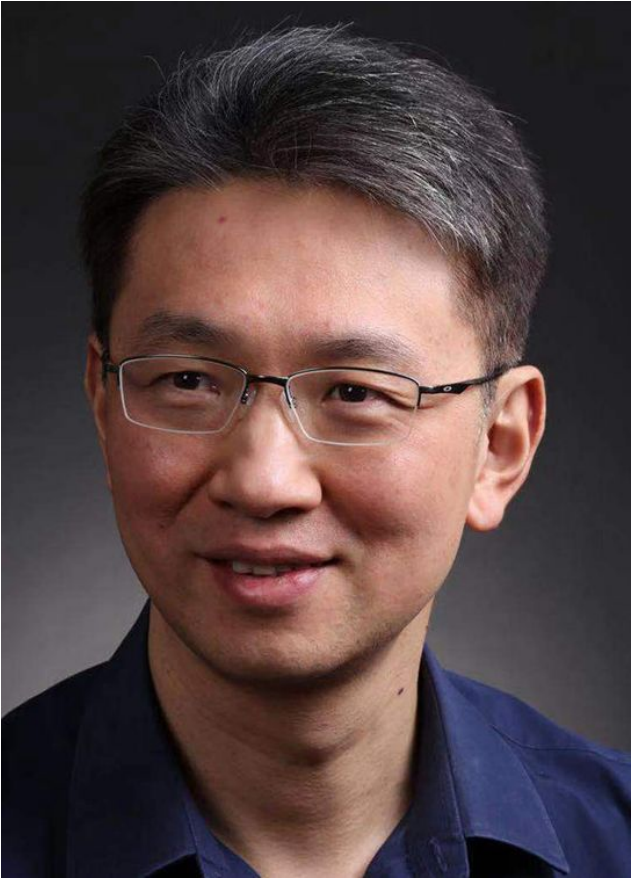}}]{Yunhao Liu} (Fellow, IEEE) received his B.E. degree from the Department of Automation, Tsinghua University, Beijing, in 1995. He received his M.A. degree from Beijing Foreign Studies University, Beijing, in 1997. He received his M.S. and Ph.D. degrees in computer science and engineering from Michigan State University, East Lansing, in 2003 and 2004, respectively. He is a professor in the Department of Automation and the dean of the Global Innovation Exchange, Tsinghua University, Beijing. He is a fellow of CCF, ACM and IEEE. His research interests include Internet of Things, wireless sensor networks, indoor localization, the Industrial Internet, and cloud computing.
\end{IEEEbiography}
\vspace{-0.5cm}

\begin{IEEEbiography}
[{\includegraphics[width=1.3in,height=1.25in,clip,keepaspectratio]{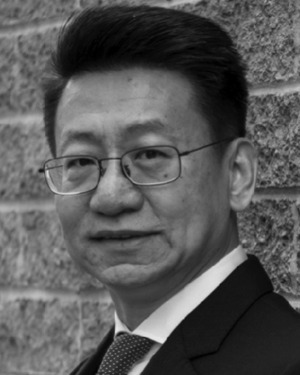}}] {Xiao-Ping Zhang} (Fellow, IEEE) received the B.S. and Ph.D. degrees in electronic engineering from Tsinghua University, Beijing, China, in 1992 and 1996, respectively. 
He is a professor with Shenzhen International Graduate School, Tsinghua University, Shenzhen, China. Since Fall 2000, he has been with the Department of Electrical, Computer and Biomedical Engineering, Ryerson University, Toronto, ON, Canada, where he is currently a Professor and the Director of the Communication and Signal Processing Applications Laboratory. In 2015 and 2017, he was a Visiting Scientist with the Research Laboratory of Electronics, Massachusetts Institute of Technology, Cambridge, MA, USA. His research interests include sensor networks and the Internet of Things (IoT), machine learning, statistical signal processing, image and multimedia content analysis, and applications in big data, finance, and marketing. Dr. Zhang is a fellow of the Canadian Academy of Engineering and the Engineering Institute of Canada. He was a recipient of the 2020 Sarwan Sahota Ryerson Distinguished Scholar Award and the Ryerson University Highest Honor for scholarly, research, and creative achievements. He was selected as the IEEE Distinguished Lecturer by the IEEE Signal Processing Society from 2020 to 2021 and the IEEE Circuits and Systems Society from 2021 to 2022.
\end{IEEEbiography}
\vspace{-0.5cm}

\begin{IEEEbiography}
[{\includegraphics[width=1.3in,height=1.25in,clip,keepaspectratio]{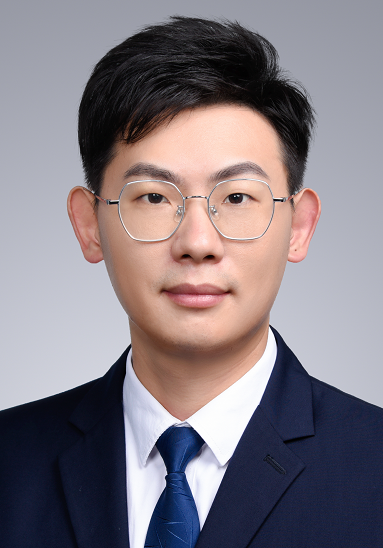}}]{Xinlei Chen} (Member, IEEE) received the B.E. and M.S. degrees in electronic engineering from Tsinghua University, China, in 2009 and 2012, respectively, and the PhD degrees in electrical engineering from Carnegie Mellon University, Pittsburgh, Pennsylvania, in 2018. 

He was a postdoctoral research associate in Electrical Engineering Department, Carnegie Mellon University, Pittsburgh, Pennsylvania. He is currently an Assistant Professor with Shenzhen International Graduate School, Tsinghua University, Shenzhen, Guangdong, China. He is also with Pengcheng Lab, Shenzhen, Guangdong, China and RISC-V International Open Source Laboratory, Shenzhen, Guangdong, China. His research interests include AIoT, artificial intelligence, pervasive computing, cyber physical system, robotics, urban sensing, brain computer interface and human computer interface.
\end{IEEEbiography}

\end{document}